\documentclass{article}

\usepackage{arxiv}
\usepackage{graphicx}
\usepackage{amsmath}
\usepackage[ruled,vlined]{algorithm2e} 

\usepackage{amssymb}
\usepackage{amsopn}
\usepackage{multirow}
\usepackage{rotating}
\usepackage{wrapfig}
\usepackage{color}
\usepackage[normalem]{ulem}
\usepackage{listings}
\usepackage{pseudocode}
\usepackage{graphicx}
\usepackage{url}
\usepackage[colorlinks=false,pdfborder={0 1 1},pagebackref=true]{hyperref}
\usepackage{indentfirst}
\usepackage{grffile}
\usepackage[utf8]{inputenc}
\usepackage{array,ragged2e}
\usepackage{dirtytalk}
\usepackage[table]{xcolor}
\usepackage{gensymb}
\usepackage{subcaption}
\usepackage{textcomp}
\usepackage{doi}

\newcommand{\excise}[1]{}

\newcommand{\ignore}[1]{}
\newcommand{\out}[1]{}
\newcommand{\OUT}[1]{}

\PassOptionsToPackage{hyphens}{url}\usepackage{hyperref}
\newcolumntype{P}[1]{>{\RaggedRight\arraybackslash}p{#1}}

\newtheorem{thm*}{Theorem}[section]

\title{
Planning Folding Motion with Simulation in the Loop Using 
Laser Forming Origami and Thermal Behaviors as an Example
}

\author{Yue Hao, Weilin Guan, Edwin A Peraza Hernandez, and Jyh-Ming Lien
\thanks{Hao and Lien are with the Department of Computer Science, George Mason University, Fairfax, VA, USA 22030.
        {\tt\small \{yhao3,jmlien\}@gmu.edu} Guan and Hernandez are with the Department of Mechanical and Aerospace Engineering, University of California, Irvine, Irvine, CA, USA 92697. 
        {\tt\small \{weiling,eperazah\}@uci.edu}}
}

\hypersetup{
pdftitle={Planning Folding Motion with Simulation in the Loop Using Laser Forming Origami and Thermal Behaviors as an Example},
pdfsubject={q-bio.NC, q-bio.QM},
pdfauthor={Yue Hao, Weilin Guan, Edwin A Peraza Hernandez, and Jyh-Ming Lien},
pdfkeywords={Laser Forming, Motion Planning,  Computational Origami},
}
\begin{document}

\maketitle

\begin{abstract}

Designing a robot or structure that can fold itself into a target shape is a process that involves challenges originated from multiple sources.
For example, the designer of rigid self-folding robots must consider foldability from geometric and kinematic aspects to avoid self-intersection and undesired deformations.
Recent works have shown success in estimating foldability of a design using robot motion planners. 
However, many foldable structures are actuated using physically coupled reactions (\emph{i.e.}, folding originated from thermal, chemical, or electromagnetic loads). Therefore, a reliable foldability analysis must consider additional constraints that resulted from these critical phenomena. 
This work investigates the idea of efficiently incorporating computationally expensive physics simulation within the folding motion planner
to provide a better estimation of the foldability. In this paper, we will use laser forming origami as an example to demonstrate the benefits of considering the 
properties beyond geometry. We show that the design produced by the proposed method can be folded more efficiently. 
\end{abstract}

\section{INTRODUCTION}



A predominate problem in design of (self-)foldable robots and structures is that no currently available design algorithm accounts for the influence of
manufacturing and material uncertainty in the robustness, such as resistance to external load and temperature
perturbations, and overall performance of self-folding structures. 
The literature of computational Origami or Kirigami considers only the generation of a valid cut pattern with simple optimization criterion, such as cut length \cite{schlickenrieder1997nets}.
To implement foldable structures in real applications, such influences of uncertainty must be quantified and accounted
for in their design to ensure that the structures retain their physical integrity regardless of the
inherent stochastic nature of real materials and manufacturing processes, and surrounding conditions \cite{hernandez2015analysisand}.

\begin{figure}[h]
    \centering
    \includegraphics[width=0.6\textwidth]{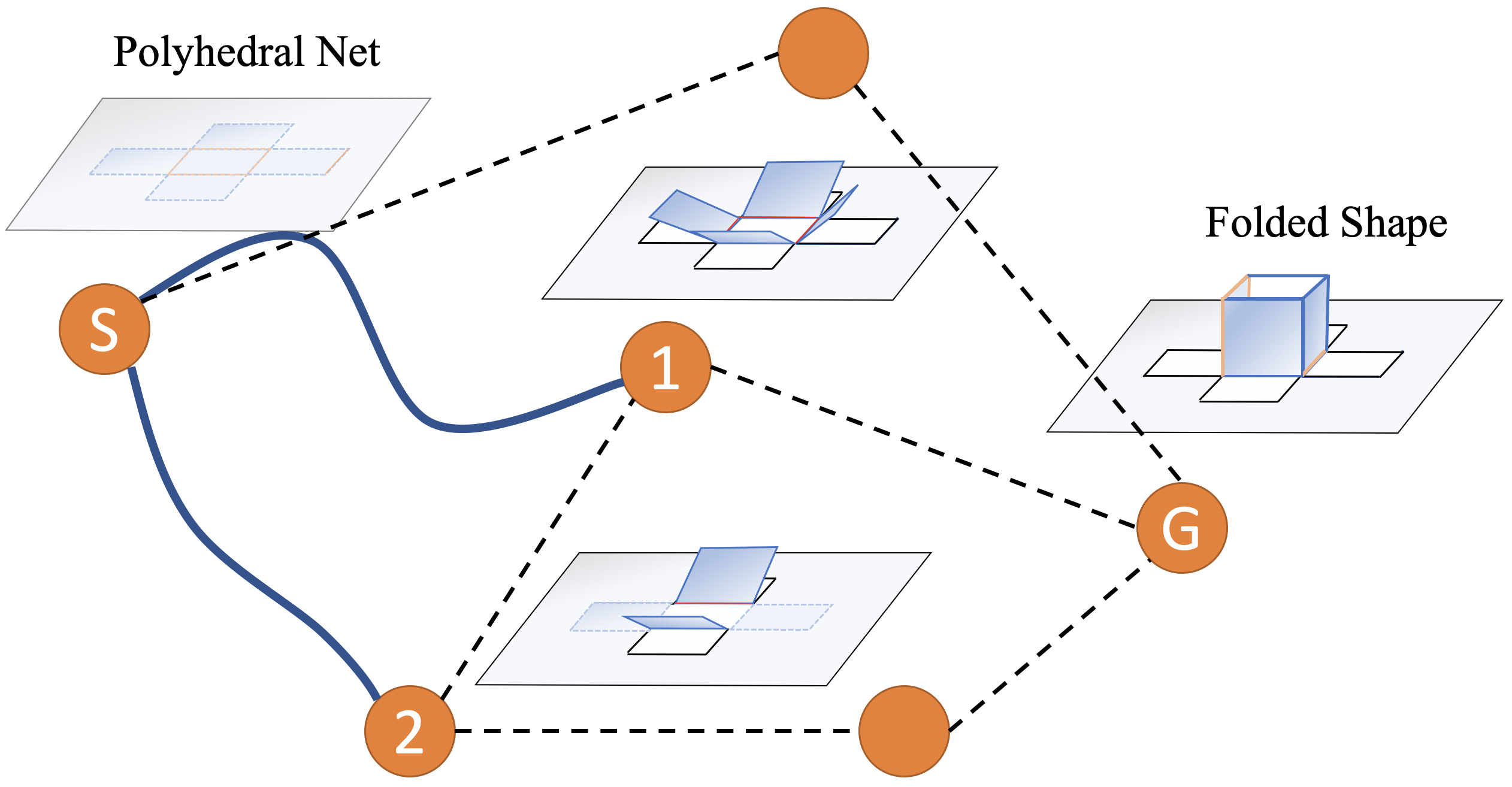}
    \caption{{Finding an optimal design in the PRM roadmap using A*: the roadmap is built based on a low-cost distance metric $d$ (dashed lines) without the thermal simulation. The search will expand the nodes with the lowest folding cost estimated via the computationally expensive distance metric $d*$  in thermal simulation (solid line) in combination with $d$ as a heuristic. 
    }}
    \label{fig:intro}
\end{figure}

\begin{figure*}[h]
\vspace*{0.1in}
    \centering
    \includegraphics[width=0.18\textwidth]{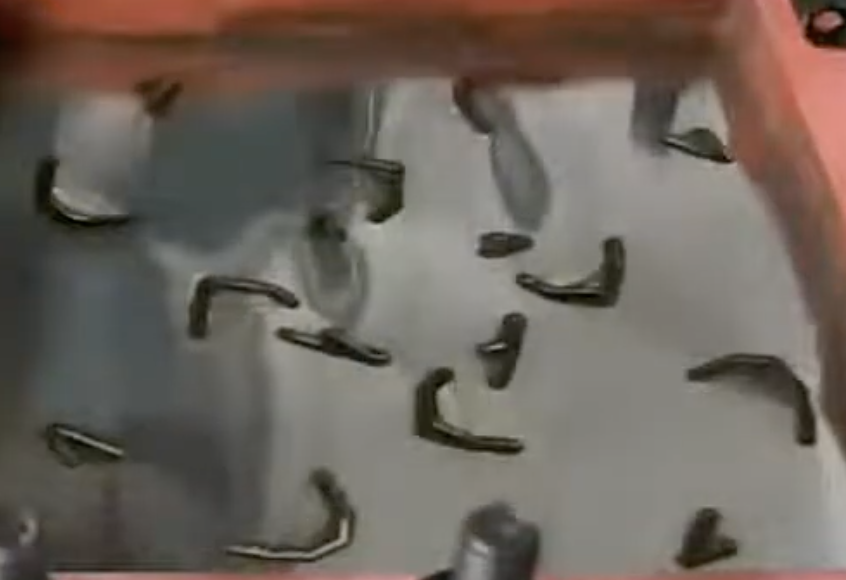}
    \includegraphics[width=0.18\textwidth]{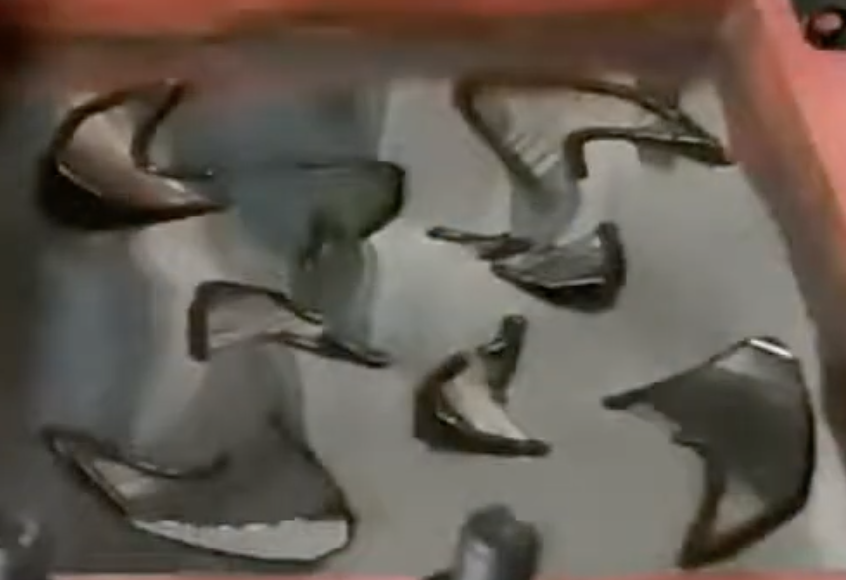}
    \includegraphics[width=0.18\textwidth]{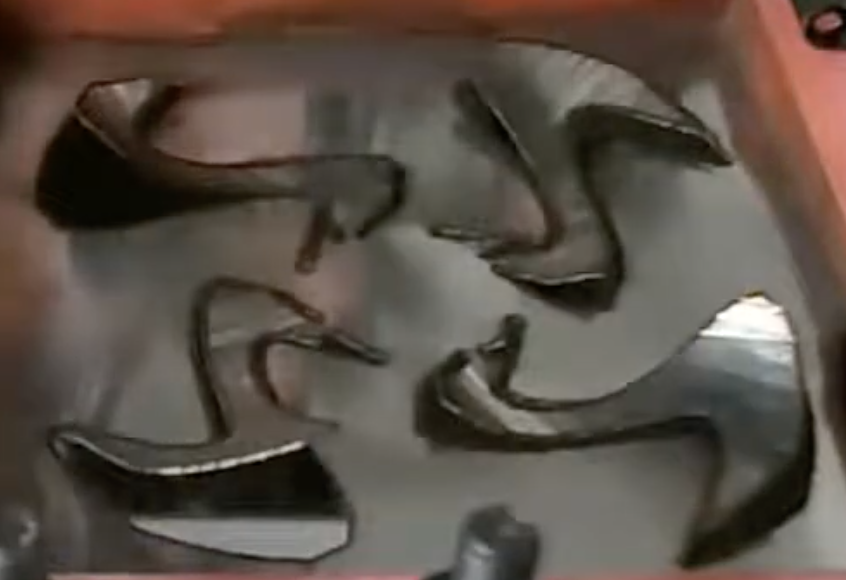}
    \includegraphics[width=0.18\textwidth]{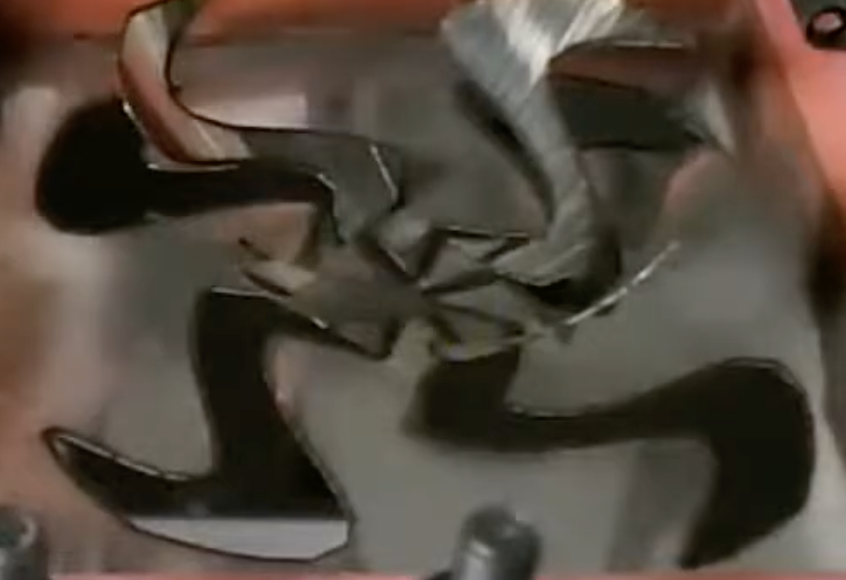}
    \includegraphics[width=0.18\textwidth]{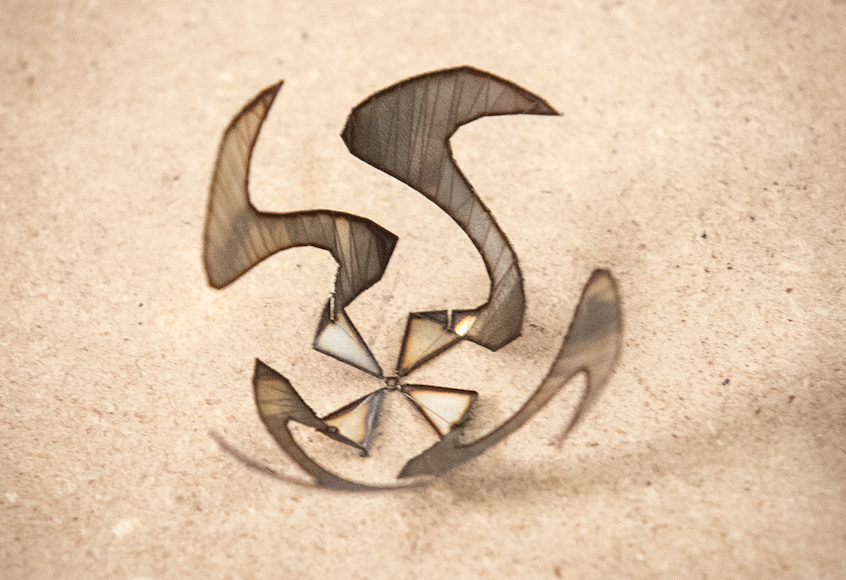}\\
    \includegraphics[width=0.18\textwidth]{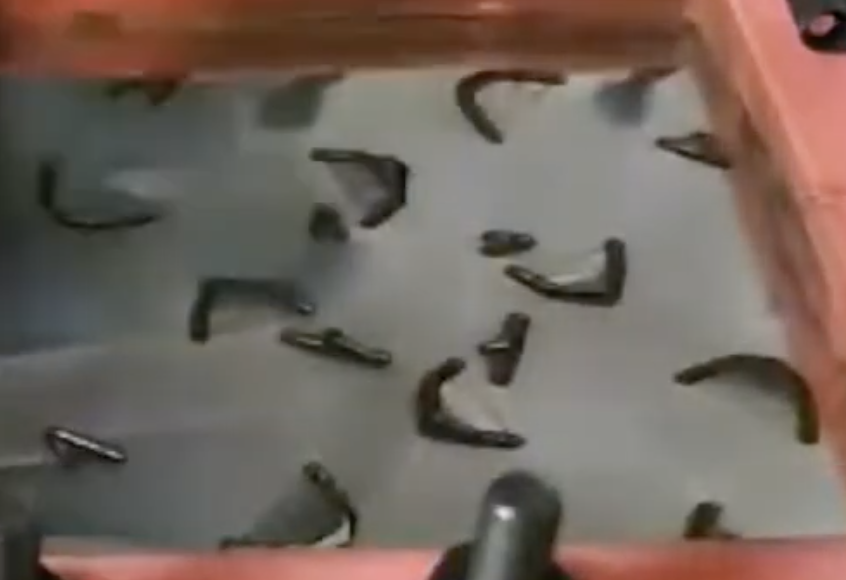}
    \includegraphics[width=0.18\textwidth]{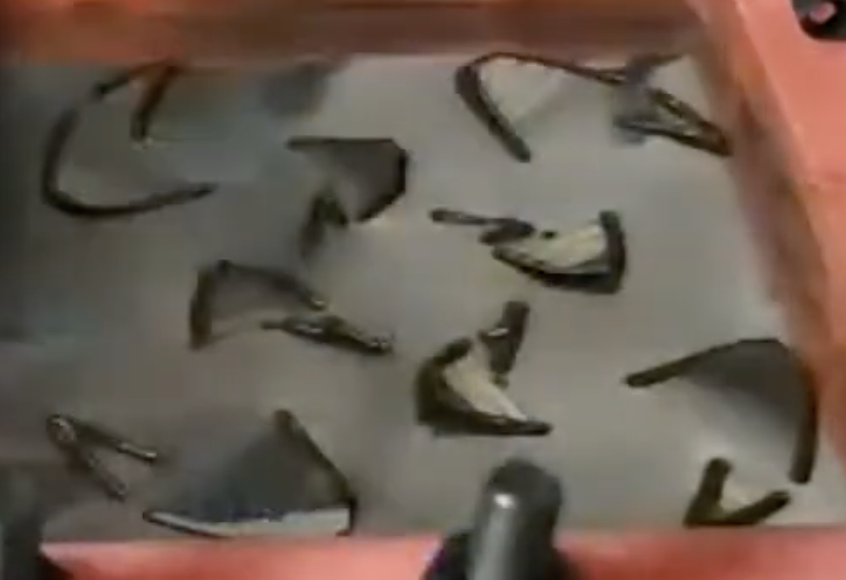}
    \includegraphics[width=0.18\textwidth]{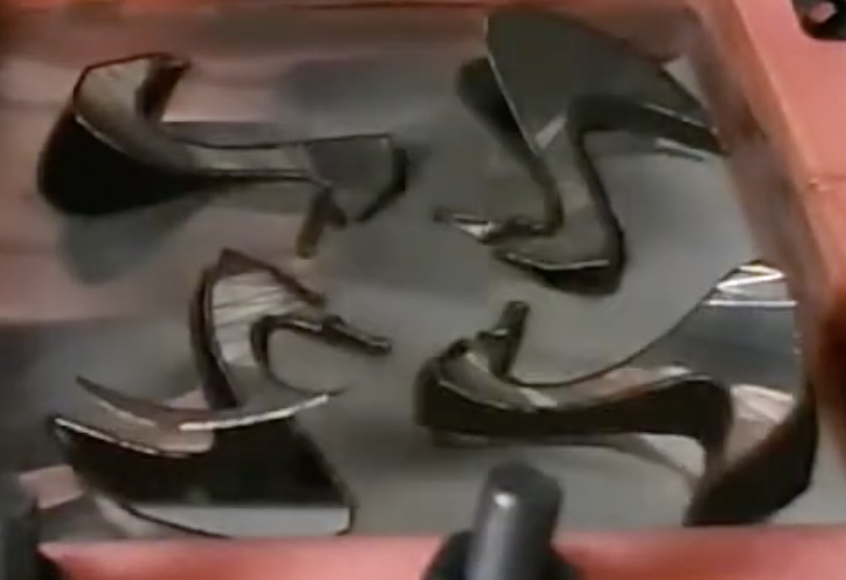}
    \includegraphics[width=0.18\textwidth]{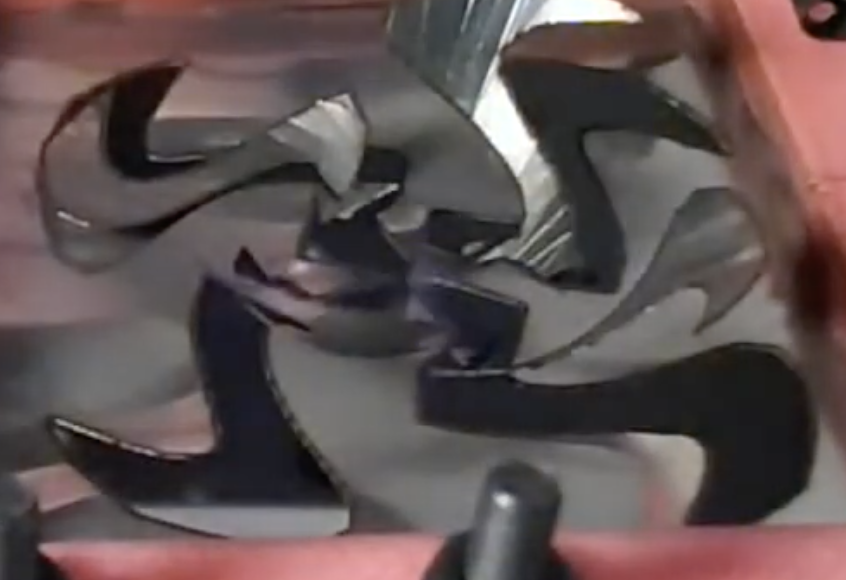}
    \includegraphics[width=0.18\textwidth]{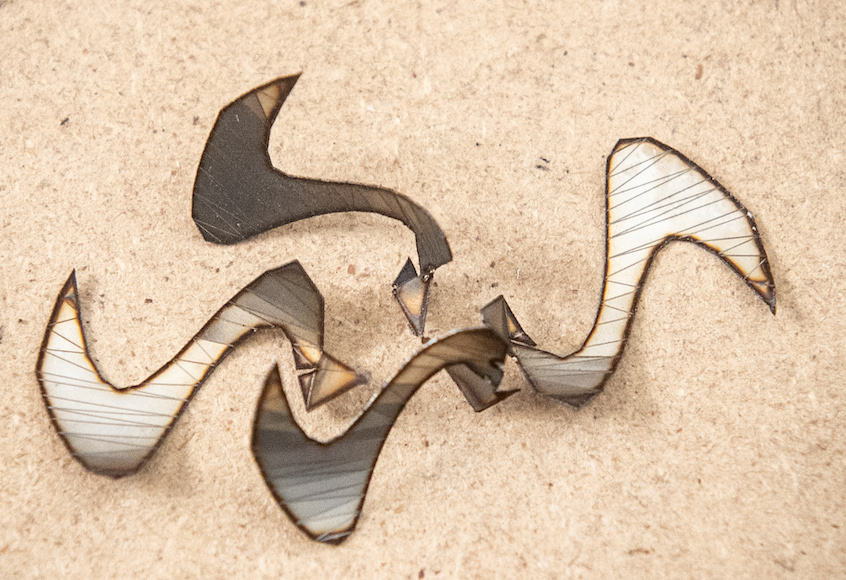}\\
    \includegraphics[width=0.18\textwidth]{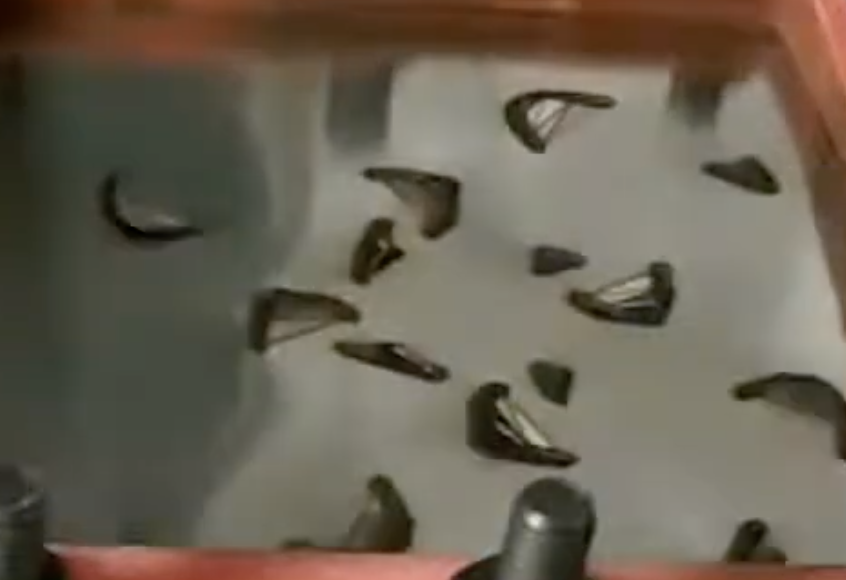}
    \includegraphics[width=0.18\textwidth]{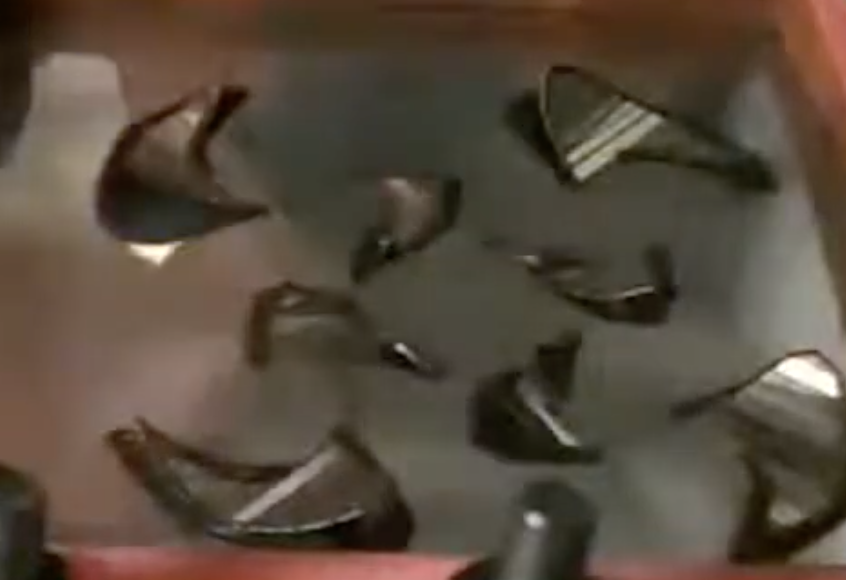}
    \includegraphics[width=0.18\textwidth]{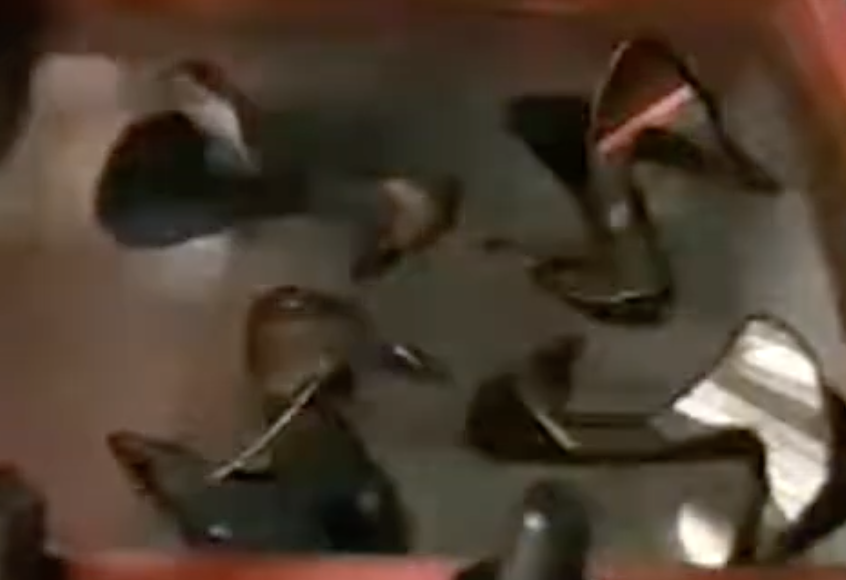}
    \includegraphics[width=0.18\textwidth]{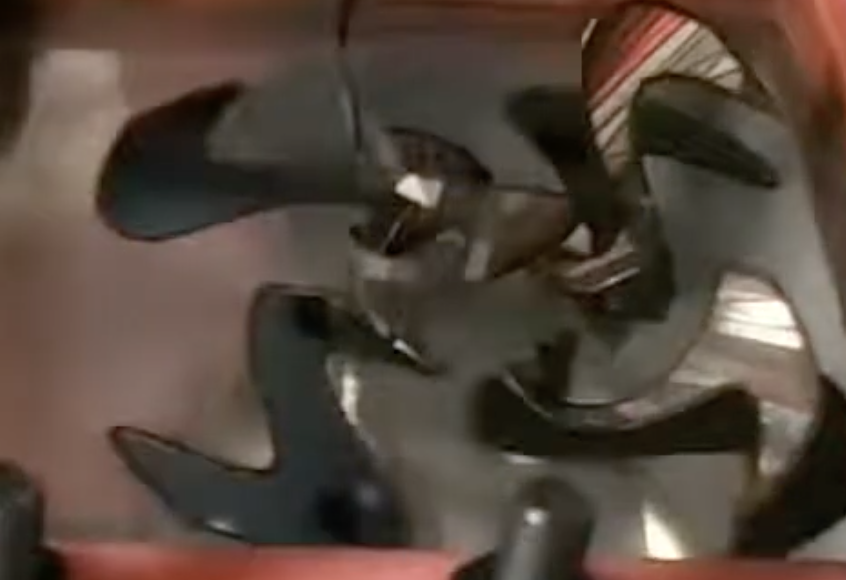}
    \includegraphics[width=0.18\textwidth]{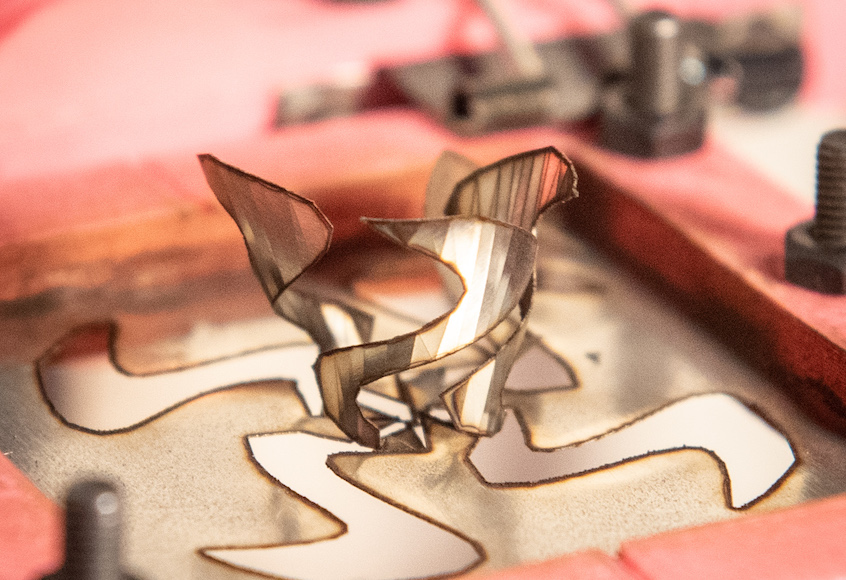}
    \caption{Laser forming the metal sheets into the sinuous antenna. A brief cooling time after each cut and fold is necessary. Using a uniform cooling delay (2 seconds), the top row shows a successful fabrication of sinuous antenna after 40 minutes. While the mid row with minimum delay can be completed in 12 minutes, it results in failure due to overheated workpiece. The buttom row shows the folding sequence and delay time generated by our motion planning algorithm, the laser forming completed in 16 minutes for a successful fabrication. Side-by-side comparison of the fabrication process is provided in the supplemented video. }
    \label{fig:frame}
\end{figure*}

Motion planners have been used {\em indirectly} in foldability optimization \cite{hao2018creat} that ensures the produced designs have simple folding motions and are free of collisions during folding.
However,  properties beyond geometric and kinematic, such as temperature, are rarely considered in the optimal design, mostly because of their formidable computational cost. Especially in the fabrication process, the workpiece constantly changes its shapes and topologies.
In addition, the successes of probabilistic motion planners heavily rely on the fact that the two intrinsically different problems, namely graph search and feasibility checking, can be decoupled. The effectiveness of these planners are limited in the cases where the two problems cannot be decoupled as in this work. 
Illustrated in Fig.~\ref{fig:intro},  the state of such robot must be determined using all configurations on the {\em unknown} path from the initial unfolded state.   

Using the computer as an analogy, a CPU needs to process a set of tasks, and certain tasks are, for example, IO heavy, while others are computationally intensive, and tasks often are interdependent. The CPU will thermal throttle if the certain tasks overheat it. Ideally, we can measure the temperature in real time and reduce the workload. However, in an open-loop system where the temperatures are not measurable, a practical solution would simulate the thermal behavior  and rearrange the task sequence which could possibly achieve a better performance. In this paper, we  use \textit{laser forming origami} as an example to explore the solution to this problem, namely foldable structure design using motion planning with physically-based simulation in the loop.


\textbf{Laser-forming origami}, as shown in Fig.~\ref{fig:frame}, uses laser to cut and fold metal sheet into 3D structure. 
Recent works have demonstrated that laser-forming origami can fabricate structures, such as antenna and inductors, faster, cheaper and more precise than 3D printed metal structures \cite{lazarus2017laser}. 


Viewing the laser-forming origami as a robotic manipulation problem has enabled computational design of complex structures \cite{hao2019comp}. 
Motion planning plays a key role in this design optimization to ensure that
the manipulator (laser beam) can deform a sheet continuously without occlusion and collision into a target shape.
However, in addition to the issue of large degrees of freedom of the joint configuration space of the manipulator and the folded structure, 
existing optimization pipeline ignores the robustness of the fabrication process and the efficiency in fabrication time. 
Examples of these issues are illustrated in Fig.~\ref{fig:frame}.



\textbf{Our Contributions}. 
This work addresses the efficiency and robustness problem in the laser forming fabrication technology by carefully integrating the motion planner and thermal simulation. We present the first physically-based simulation of laser forming, and based on which we designed a simplified modeling of the thermal behavior of workpiece with dynamic typologies. We then incorporate such simulation in the motion planner to find a folding motion for the fabrication process to prevent the thermal damages. Unlike  existing algorithms that simply uses simulation in the local planner, we propose an A$^*$ based searching strategy that traverses the roadmap efficiently and obtain the optimal path.

Fig.~\ref{fig:intro} illustrates the need of such a planner. 
Consider the configuration $\#1$ that may appear to be a desirable configuration in finding the shortest path for folding the open cube model. However, the cutting four sides at the same time will generate intensive heat flux and time delay is required to avoid over heating, which may undesirably deform the structure. On the other hand, via configuration $\#2$, though takes two steps to fold, is actually faster in fabrication as the thermal condition is better controlled with lesser cutting happening at the same time.

\section{BACKGROUND AND RELATED WORK}


\subsection{Simulation-in-the-loop Path Planning}

The use of physically-based  simulation in a path planning framework is not uncommon. It has been shown that, in contrast to the traditional feedback systems that require linearization of the state transfer function, simulation-in-the-loop planners can learn accurate physical parameters and result in more robust controls navigating robot through difficult terrains \cite{keivan2013realtime2} or even highly deformable environments \cite{rla-pmcde-06}. 
In most, if not all, of these simulation-in-the-loop planners, simulations are used to provide feed-forward control in order to expand the stochastic search in the state space, such as RRT-based methods \cite{lk-rkp-01}. In this work, thermal dispersion and the folding motion induce by the thermal stresses (see details below) are simulated to not only for the purpose of finding valid folding motion but also for ensuring the quality and reproducibility of the planned motion.  

\subsection{Laser Origami and Laser Forming}

Laser origami uses a laser source to fold 2D sheets into 3D structures. Several folding mechanisms have been proposed, including polymer softening \cite{mueller2013laser}, release of pre-stressed films \cite{alberto2012laser} and laser forming \cite{shen2009modelling}, in which, laser heats up the workpiece to create plastic thermal stresses for controlled folding \cite{paramasivan2018experimental}. 

Several distinct modes of laser forming are possible, including the temperature gradient mechanism (TGM) and buckling mechanism (BM). In TGM bending, the laser is rapidly scanned across the surface, creating a vertical temperature gradient through the metal thickness.  Initially the top expands relative to the bottom (``counterbending'') \cite{madsond2014development}; however, the cooler surroundings push back against this expansion, resulting in compressive plastic strains.  When the workpiece is subsequently allowed to cool, the initial expansion disappears but the compressive plastic strains remain in the material, resulting in bending back upward toward the laser source \cite{shen2009modelling}.  A finite element-based simulation of TGM is presented in Appendix A.

For the buckling mechanism, the laser is scanned more slowly to allow heat to reach all the way through the thickness of the workpiece.  The heated region again expands, generating compressive plastic stresses as before, but this time the compressive stresses are against the full thickness of the heated plate.  Compressive stresses in a thin membrane result in buckling of the heated region \cite{edwardson2016astudy} , and as the laser is scanned the buckling results in the folding of the workpiece.  Unlike the TGM mode, the buckling mechanism can cause both upward or downward folding, and it is necessary to control the direction through an additional force such as a pre-strain in the material  \cite{lazarus2017laser}. 
Recently, a method combining laser cutting and laser forming can fold structures with both valley and mountain folds, using  TG and buckling mechanisms, respectively,  from a metal sheet  \cite{lazarus2017laser}.  
In theory,  complicated structures are possible using this technology; however, in practice laser origami has been limited to simple designs and manual placement of folds.  

\subsection{Foldable Structure Optimization}
\label{sec:structure}
Heuristic methods \cite{schlickenrieder1997nets,prautzsch2005creating} have been developed to unfold convex polyhedra. Finding nets of non-convex shapes is  significantly more challenging and segmentation is often needed to avoid overlapping \cite{prautzsch2005creating,takahashi2011optimized,xkkl-smi-16}.
For (self-)foldable structures, collision-free folding motion that brings it back to the 3D shape is essential but non-trivial.
The motion of these nets is typically found by a motion planner that models the net as a tree-like articulated robot \cite{song2004motion,xl-iros-15}.
However, depending on the polyhedral nets, the planning can take hours to find a path and ends up with complex motions that are not desirable for physical realization. 
Until recently, Hao et al.  \cite{hao2018creat} proposed an optimization framework to optimized the foldability using the geometry and topological attributes of the net.
The resulting nets have significantly simpler folding motion than an arbitrary net does.

\section{LASER FORMING ORIGAMI MODELING}
\label{sec:prelim}


\subsection{Laser Forming Modeling}

\begin{figure}[h]
    \centering
        \begin{subfigure}[b]{0.3\textwidth}
        \includegraphics[width=\textwidth]{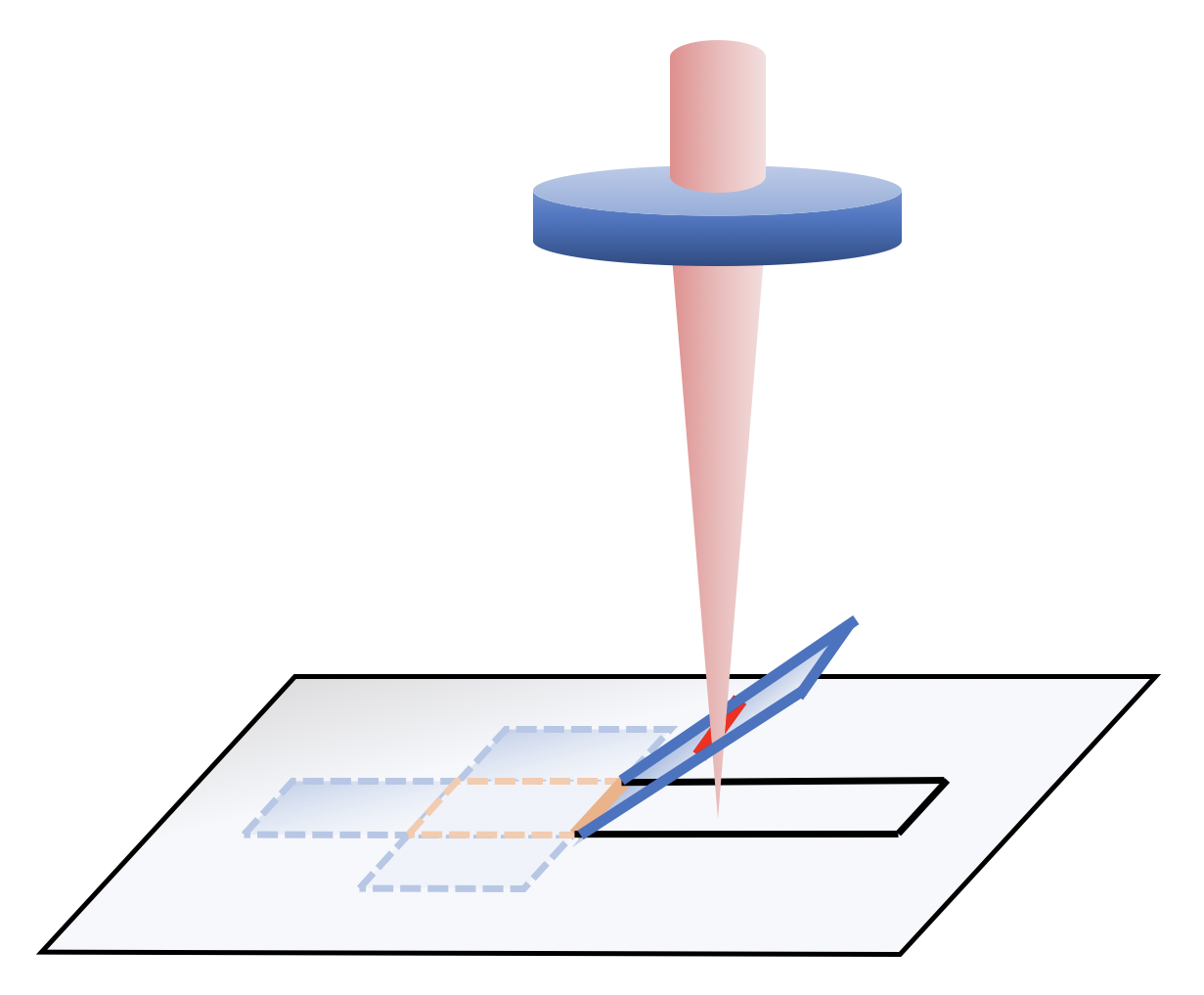}
        \caption{}
        \end{subfigure}
        \begin{subfigure}[b]{0.3\textwidth}
        \includegraphics[width=\textwidth]{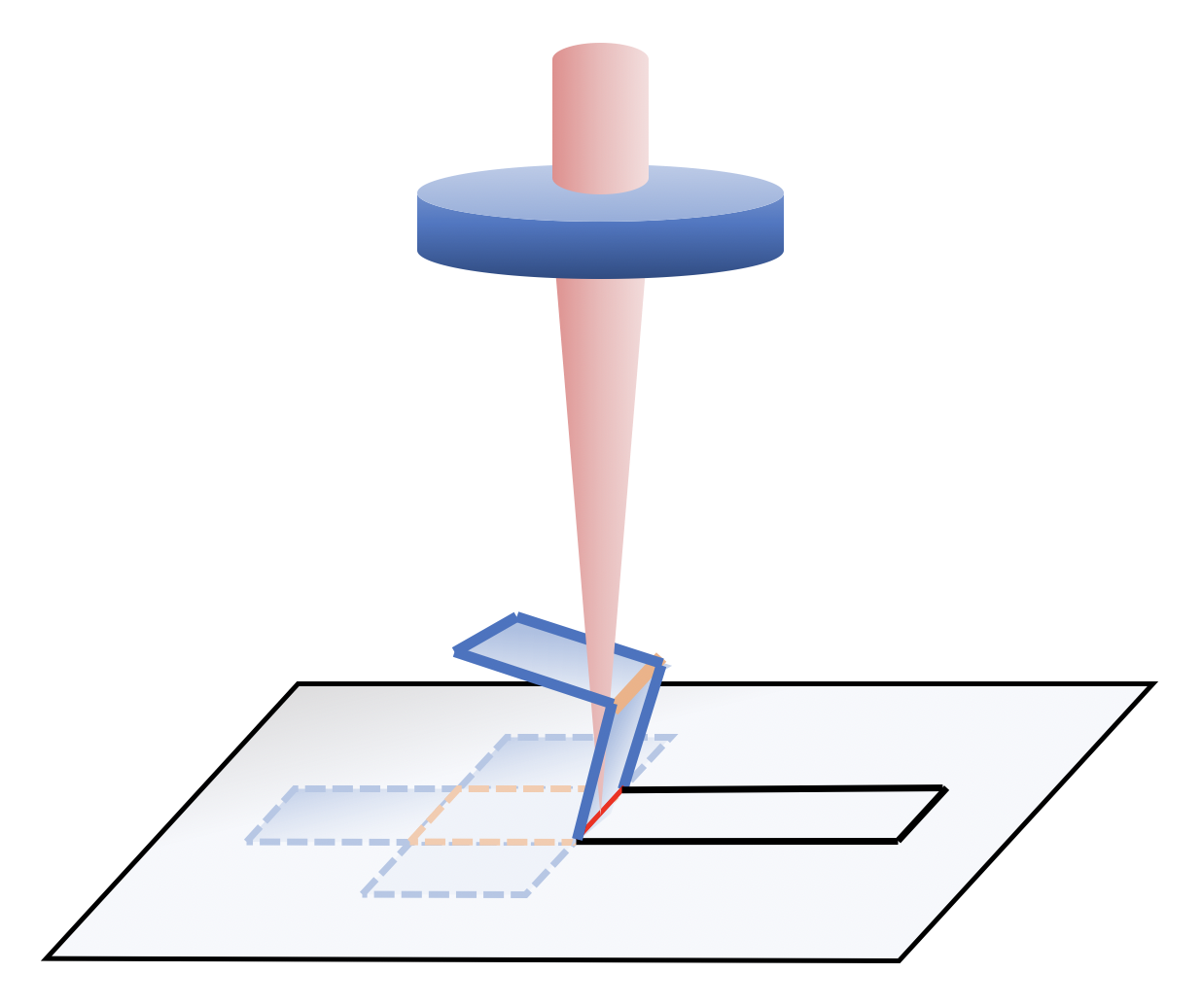}
        \caption{}
        \end{subfigure}
    \caption{An illustration of the geometric constraints of laser forming. (a) The workplane constraints: the laser can only focus on the 2D plane of the substrate through the lens. The crease (shown in red) can not be folded because it is out of focus after being lifted. (b) visibility constraints: the folding crease must be visible from the laser source. The crease (shown in red) cannot be folded further because the laser is blocked by the folded part.}
    \label{fig:setup}
\end{figure}

A commercially available fiber laser marker with a single laser head is used in this work. Fiber laser makers and cutters typically use galvanometers to reflect and move the laser beam on in $x,y$ directions, situated above the substrate. Shown in Fig. \ref{fig:setup}, the substrate is cut into a single connected piece by laser passing along the edge of the polyhedral net \cite{schlickenrieder1997nets}, which is a 2D unfolding of the target shape. Due to the optical focal length of the lens is fixed, the laser beam should only intersect with the metal substrate on a 2D work plane $z=0$, bounded by the working area, otherwise the intersection is out of focus with weakened strength which results in undesired folding or deformation. As the intersection point $p = (x,y,0)$ move along the edges of the polygonal edges of the target shape in the work plane, it causes the metal to be cut and fold into 3D with different power and speed settings $w$. The laser machine emits laser with three different power strengths based on the material properties of the substrate, strong power for cut, weak for fold and 0 for the delay. In our work, we use a $100um$ stainless steel, which needs 20 watts to cut and 2 watts to fold. The delay is necessary because the laser may cause extreme heating during cutting and folding and time delay must be allowed to prevent the substrate from being deformed or damaged by overheating. The problem of laser forming ultimately seek to find the motion path for the intersection point $p=(x,y,w)$ to cut and fold a polyhedral net into a polyhedral mesh. However, planning the motion in the $p=(x,y,w)$ space is non-trivial, because the folding of the polyhedral net is much more complex to due to workplane and visibility constraints (Fig.~\ref{fig:setup}) that the folding process must follow. 
Therefore, the motion planning is done on the folding process of the sheet itself in order to prevent the self-intersection of the material. Consider the polyhedral net a set of links (polygons) connected in a tree structure, $L = \{l^i\}$, and the folding motion is a set of configurations of those links along time $t$, $Q(t) = \{q^i_t\}$.  The laser forming process is a cumulative effect of multiple laser passes applied on the same crease.  On each laser pass, the crease only folds a small degree. The number of laser passes versus the folding angle can be determined via experiments for a particular hardware and material setup~\cite{lazarus2017laser}. Once the folding motion of the sheet $Q(t)$ is obtained, the motion of $p$ can be easily obtained as $p$ simply moves at constant speeds for cutting and folding respectively.

\subsection{Thermal Constraints}
\label{sec:laser_forming}

In addition to the geometric constraints described above, because laser forming uses thermal plastic stresses to bend a crease, the effect of heat being distributed across the sheet is critical to the efficiency and quality of the fabrication results.
We should allow time delay between cuts and folds to ensure that the folded structure and the sheet are free of undesired deformation (\emph{e.g.}, bulging).
Existing solutions either do not consider the thermal constraints and simply use a conservative long delay between laser passes~\cite{lazarus2017laser}. Or have an adaptive delay time based on the estimation of the substrate temperature during the fabrication process. For example, the fabrication will pause if certain laser pass will result in over-heating. Both approaches will result in inefficiency in fabrication as time has been wasted for delays.
The primary objective of this work is based on the simple fact that if the folding motion can be optimized with the consideration of the thermal constants in the planning stage, then hours can be saved laser in the fabrication (for instance, laser folding two areas that are far apart being folded simultaneously without causing overheating due the laser passes are less concentrated). Making this technology a more attractive approach in terms of efficiency and quality in comparison with the traditional fabrication process such as 3D printing.
The challenge mainly comes from two aspects. Firstly, having the thermal constraints in the motion planning is simply more complex and such simulation cannot be simple integrated in the motion planner using any of the existing algorithms. We will have a detailed analysis of such in Section \ref{sec:optimal}, to show how the thermal planning and the motion planning forms circular dependencies which make the planning difficult.
Secondly, it is computationally expensive to simulate the heat transfer of the substrate that constantly changing the geometry as being cut and fold. 
A reliable thermal simulation using finite element analysis software takes a long time even without changing the connectivity of the components. Therefore, to be discussed in Section \ref{sec:heat} our second contribution is to obtain a relatively realistic heat transfer modeling that is also fast enough to be integrated into the motion planner.
However, thermal field data from experimentally calibrated high-fidelity simulations can be potential integrated into the motion planner to obtain more physically accurate results.  Appendix A describes our preliminary method based on employing finite element analysis to predict the deformation and temperature fields experienced by the sheet during laser folding. It includes information on potential calibration routes and implementation.

\section{MOTION PLANNING WITH HEAT SIMULATION IN THE LOOP}


\subsection{The Optimal Motion Planning Problem}
\label{sec:optimal}
 In the outside-in algorithm, a PRM planner will output the $Q(t)$ which can be executed smallest cost $c$.
Our objective is to find an optimal laser trajectory that minimizes the folding time while satisfies the thermal, visibility and foldability constraints.
First, we know that $l^i$ must be cut before folding, then we can consider a cutting plan $C(t) = \{p^i_t\}$ for the time of certain link $i_i$ being cut by laser. As workpiece is heated by the laser cutting, we can calculate the temperature of the link along time $t$, $H(t) = \{h^i_t \}$. Both $C(t)$ and $Q(t)$ will increase the temperature on the affect region, and meanwhile $H(t)$ will decrease along time as the heat dissipate to the environment.

In summary, the folding motion $Q(t)$, cutting plan $C(t)$, and thermal state $H(t)$ forms circular dependencies, folding affects cutting, cutting affects the heat and the folding motion is constraint by the heat. Using the $\rightarrow$ to denote the effects of these three elements to each other, we have 
\begin{eqnarray}
Q(t)\rightarrow C(t)\\
C(t), Q(t) \rightarrow H(t)\\
H(t) \rightarrow Q(t)
\end{eqnarray}
In the existing literature, the research for laser forming motion planning has been focus on the $Q(t)$, simple rules are used to decouple $C(t)$ and $H(t)$ from the motion planning. For example, the easiest way to decouple $Q(t)\rightarrow C(t)$ is to cut the entire workpiece before the folding process. In this approach, the workpiece is intensively heated by the cutting laser in the beginning which causes deformation, and the folding process is unstable due to lack of support for the structure. For $C(t), Q(t) \rightarrow H(t)$, the naive method is to add a delay (usually 1-2 seconds) between each cutting/folding laser scan to prevent the damage cause by overheat. However, due to lack of planning, the delay is added even if the laser works on different regions between consecutive scans, and therefore folding process is over-delayed which results in a longer time cost for folding, \emph{i.e.,} $c^* > c$. 

In this work, we aim at incorporate $C(t), H(t)$ into the planning of $Q(t)$ in order to optimize the delay time and decrease the cost $c^*$. We propose the following rules for Eq.(1) and Eq.(3):

\textbf{$Q(t)\rightarrow C(t)$}: we want to cut $l_i$ as late as possible. Therefore, $C(t)$ is simply a function of $Q(t)$. 

\textbf{$H(t) \rightarrow Q(t)$}: In order to prevent over-heating and potential damage to the workpiece, we must add delay in the folding process. We pause the cutting and folding on link $l^i$ at time $t$ if $h^i_t > T_u$, and resume if link $h^i_t < T_l$, where $T_l < T_u$ and $T_u$, $T_l$ are constant thresholds.

For $C(t), Q(t) \rightarrow H(t)$, it has to be done by thermal simulation which computationally expensive. We proposed a simplified model derived from the finite element analysis of thermal simulation which we will discuss in section B.
Finally, we will discuss our motion planner with $H(t)$ to find the optimal $Q(t)$ in Section C.

\subsection{Heat Transfer Modeling}
\label{sec:heat}

This section discusses how the heat transfers on the substrate over time $H(t)$, and how to calculate it efficiently. In the study of heat transfer theory, the heat transfer model is typically computed based on the finite element analysis~\cite{incropera2006fund}, where the solid object is discretized onto smaller nodes and numerical solution can be obtained through simulation software. However, such simulation is often very costly for laser forming as the topology of the workpiece is constantly changing in the cutting and folding process. We use a similar method to obtain a simplified finite element estimation of the temperature on the surface of the substrate. 
It is acknowledged that this modeling approach may be inaccurate and oversimplified. 
However, our goal is to show that with a conservative estimation, \emph{i.e.,} the estimated thermal conditions are worse than reality, our algorithm can still optimize the folding path so that the fabrication efficiency can be obtained.

\begin{figure}[h]
    \centering
    \includegraphics[width=0.7\textwidth]{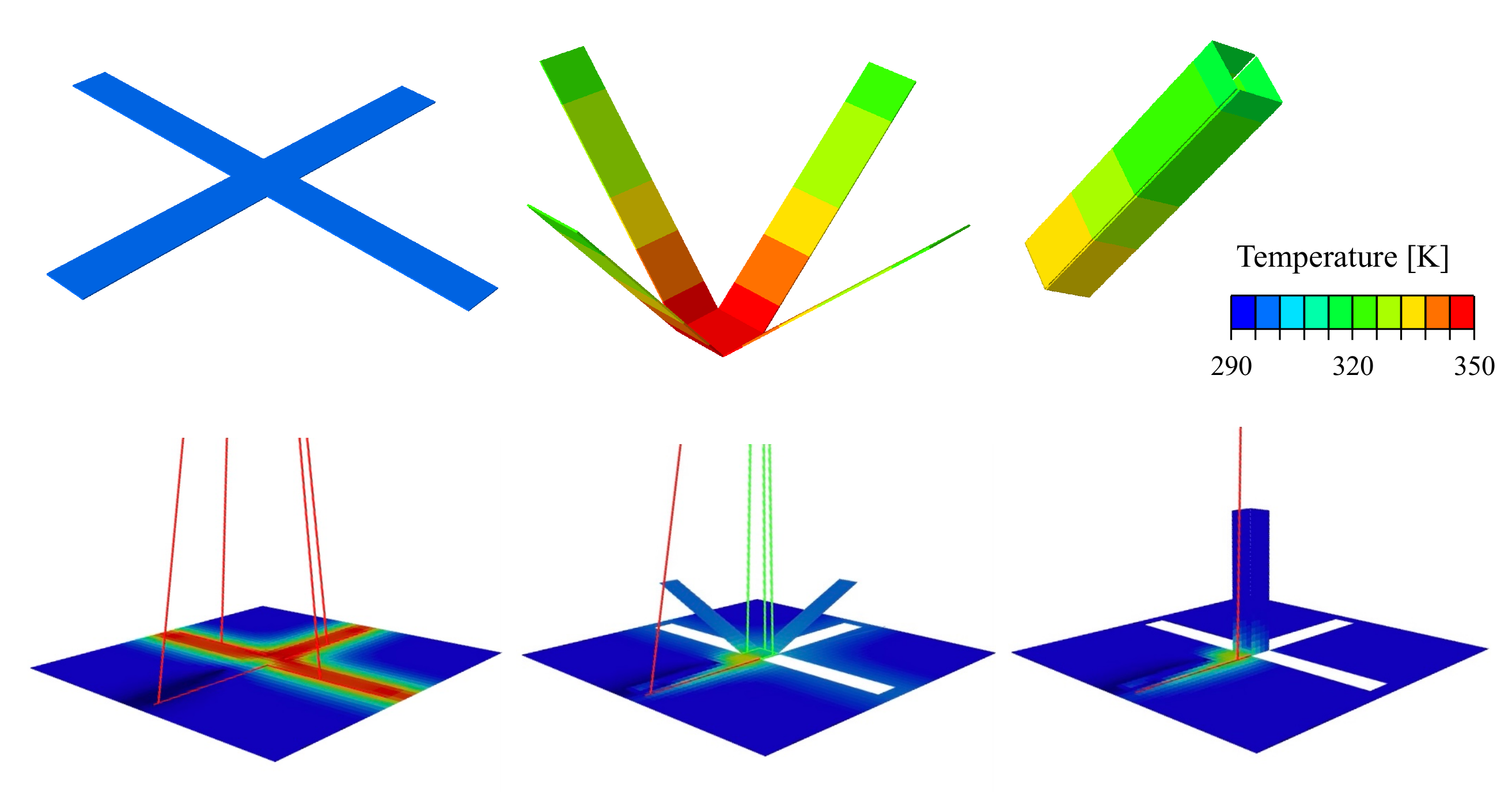}
    \caption{Top: Cross model laser forming simulated in Abaqus using the approach described in Appendix A; Bottom: Cross model laser forming simulated using proposed heat transfer model. Cutting lasers are in red and folding lasers green.}
    \label{fig:heat}
\end{figure}

In our modeling, the substrate surface is discretized into a 2D grid, where the node temperature $T_{m,n}$ at coordinate $(m,n)$ can be affected by the heating of the laser or conduction of the material. We use the finite-difference equation \cite{incropera2006fund} to update the temperature after conduction:
\begin{eqnarray}
T_{m,n+1} + T_{m,n-1}+ T_{m+1,n} +T_{m-1,n}-4T_{m,n} = 0
\end{eqnarray}
It requires simply that the temperature of a node be equal to the average of the temperatures of the four connected neighboring nodes. 
During the laser forming process, the heat flux will increase the temperature on the nodes corresponding to folding and cutting creases shown in Fig. \ref{fig:heat}. As the boundaries of the polyhedral net are cut by the laser, the connection of the corresponding nodes are also disconnected, and therefore the heat cannot transfer between these nodes. Meanwhile, the nodes are heated by the laser and dissipate the heat to the environment (see Appendix A). Fig. \ref{fig:heat} shows the simulated thermal images using this heat transfer model.

The maximum temperature of all the nodes must be controlled under a certain threshold to reduce the risk of the damage or undesired deformation of the substrate.

\subsection{Folding Path Planning}

We now discuss a PRM based motion planner \cite{xl-iros-15,song2004motion} to determine the optimal path in the configuration space, folding from the flat state to the folded state with minimum cost $c^*$.

The difficulty of solving such a motion planning problem lies in the computation of the distance metric. 
Let the distance from configuration $Q$ to $Q'$ be $d$. $d$ can be derived using efficiently using outside-in folding \cite{hao2019comp}. Note, $d$ is not reversible as laser forming is a monotonic process and therefore if $d$ for $Q$ to $Q'$ exists, then $Q'$ to $Q$ is infinitely large. However, due to the thermal constraints, $d$ also depends on the initial temperature of $Q$, \emph{i.e.,} if the path to $Q$ results in a higher temperature, $Q$ may take a longer distance to reach $Q'$.  
A naive approach to incorporate $H(t)$ to PRM is to extend the state space be $Q \times H$. However, this means significantly more difficulties in sampling because the valid samples for the temperature are rare to find. Alternatively, the $H(t)$ is treated as the environment and the thermal constraints are enforced along in the local planner alone with collision checking. However, unlike the collision constraints which is entirely dependent on the configuration of the samples. $H(t)$ can be considered the dynamics of the process and the temperature of $Q(t)$ is determined by the entire path from $t = 0$.  Therefore, unlike the collision checking problem, where we can find the shortest paths in the roadmap and greedily only check collisions on this path if the local planner is too costly to run, the shortest path with $H(t)$ involved requires running local planner on the entire roadmap, which is extremely inefficient. From this perspective, the $H(t)$ is a unique type of variable that cannot be simply combined into the sample space nor the configuration space. Here we propose a new algorithm to solve this problem.

Shown in Algorithm \ref{alg:mp}, let $d^*$ be the distance metric with thermal simulation. We sample the configurations $Q$ that satisfy the outside-in algorithm. The optimal solution is to find the shortest path from the initial configuration to the goal configuration based on distance metric $d^*$. However, based on previous discussion, $d^*$ can be obtained using finite element based simulation which is costly to compute. It is infeasible to build a roadmap $G$ in C-space based on such distance metric $d^*$. Instead, the roadmap is built based on distance $d$, and we incorporate the simulation when traversing $G$.
\begin{algorithm}[h]
    \SetKwFunction{NBU}{ThermalLaserMotionPlanner}%
    \SetKwFunction{LP}{ThermalSimLocalPlanner}%
    \SetKwProg{Fn}{def}{\string:}{}
    \caption{Thermal Laser Motion Planning}
    \label{alg:mp}
    
   \Fn(){\NBU{$L$}}{
    \KwData{Laser-formable net $L$}
    \KwResult{Optimal Path $\pi$}
    Let $Q_I,Q_G$ be the initial and goal configurations\;
    Let $H_I$ be the initial (room) temperature\;
    Let $d$ be the distance metric by out-side folding\cite{hao2019comp}\;
    Build a roadmap $G$ using PRM with $K$ nearest neighbor using $d$\;
    Let $S = {(Q_I,H_I,0,v)}$ be the set of front nodes in $G$, starting with the initial configuration and temperature, a cost of 0, and heuristic value of $v$ which is calculated from $Q_I$ to $Q_G$ using distance metric $d$\;
    Let $P[n]$ be the parent node of $n$\;
	    \While{$S$ is not empty}{
	        $n = (Q_s, H_s, c, v) $ the node in $S$ with lowest $c+v$\;
	        Remove $n$ from $S$\;
	        \For{Each node $m = (Q_e)$ in neighbor of $n$ in G:} {
	            $w, H_e = $ \LP{$Q_s$,$Q_e$,$H_s$}\;
	            $c' = c + w$ \;
	            \If{ $c'$ is lower than existing cost of $m$ } {
	                $v' = $ the distance from $Q_s$ to $Q_G$ using distance metric $d$\;
	                Save $H_e, c', v'$ to $m$ in $G$\;
	                Record $n$ being the parent of $m$ in $P$\;
	                Add $m$ to $S$\;
	            }
	        }
	        \If{ $Q_G$ is reached } {
	       Traverse $P$ to obtain $\pi$;
	        }
	    }
    }
\end{algorithm}
To further reduce the search space, we connected each node to $K$ closest neighbors using the distance metrics of $d$. A naive approach is to find the shortest path $\pi$ in $G$ (the edge weights in the graph are also estimated using $d$). For each segment in $\pi$, run the local planner with $H$ using a simulation-based local planner where collisions are checked and constraints are enforced. However, such approach only guaranteed the shortest path in distance metric $d$.
Similar to $A^*$, for each node $n$ in $G$ we store values of cost and heuristics. 
The cost value of is calculated based on distance metric $d^*$ get to the node $n$ from the start node (\emph{i.e.,} the flattened structure) and includes both the folding time and the time delay needed along the path. Then, the heuristics value of $n$ is the approximated folding time without time delay along the path from $n$ to the target node (\emph{i.e.,} the fully folded structure) using the edge weights estimated using outside-in distance metrics $d$. In addition to configuration $Q$ represented in each node, we also store the temperature $H$ in the node.
Because $d\leq d^*$, the heuristics is admissible. This guarantees that our $A^*$ search algorithm will find the optimal path.

\section{Experimental Results}

\subsection{Simulation Results}
\label{sec:sim}
Table \ref{tb:mp} shows the comparison of the effectiveness of the proposed algorithm in simulation results. We call the folding path planned using the proposed algorithm Adaptive.  The path length is total folding angle for the trajectory in rad in distance metric $d*$. Because the folding speed is a constant, the cutting and delay (time) cost is also converted to rad.
The average temperature is the average over time of the maximum temperature on the substrate surface. An interesting result is that the average temperature for our adaptive folding path is slightly higher than no-delay folding. This is because the most heat concentrates on the last few cuts near the center, which results in a high risk of overheating. The motion planner decides to choose a folding path that scarifies the early cuts where the heat is less concentrated, but reduces the maximum temperature in the later cuts and folds, which results in an overall less extreme thermal conditions but slightly higher average. Our fabrication results in Section \ref{sec:fab} also demonstrate similar behaviors.
For the computational cost comparison, we show the path planning time of the algorithms implemented in single-threaded C++ on a MacBook Pro with a 3.2 GHz Intel Core i7 CPU with 16GB Memory. Since the Sinuous Antenna model is a fairly simple model and it is straightforward to find a valid folding path, we only use a sampling size of 20 and $K = 10$ nearest neighbor to build the roadmap. 
For uniform delay or no-delay path, the planner only checks self-collisions and no thermal simulations are involved to verify the path for rigidity and visibility constraints. 
Our thermal simulation has comparable time cost as the collision checking in the motion planner for this specific model. 
However, we found that increasing the thermal simulation resolution, \emph{i.e.,} the grid size which currently $50\times50$, will significantly increase the computational cost.

\begin{table}[h]
\caption{Planning and Simulation Results for Laser Forming Sinuous Antenna}
\centering
\begin{tabular}{|l|c|c|c|}
 \hline
  & Uniform & No Delay & Adaptive \\ \hline
 Path Length (rad*) & 43.42 & 10.93 & 13.37 \\ \hline
 Average Temperature (\textcelsius{}) & 38.02 & 61.42 & 72.10 \\ \hline
 Planning Time (s) & 141.2 & 102.3 & 175.3 \\ \hline
\end{tabular}
\label{tb:mp}
\end{table}

\begin{figure}[h]
    \centering
    \includegraphics[trim=0 0 0 130,clip,width=0.2\textwidth]{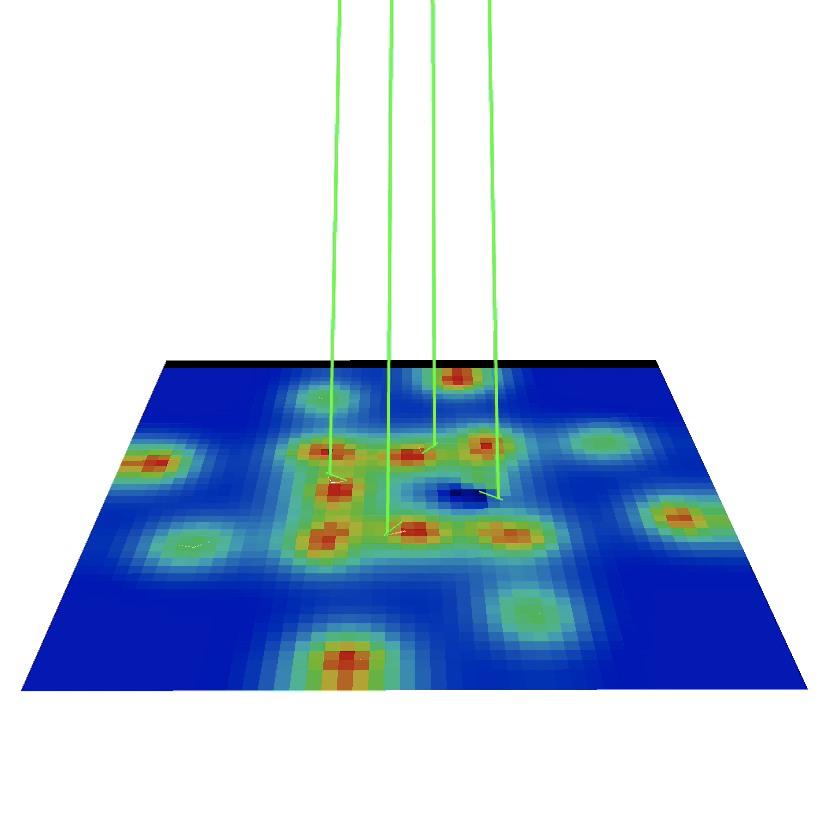}
    \includegraphics[trim=0 0 0 130,clip,width=0.2\textwidth]{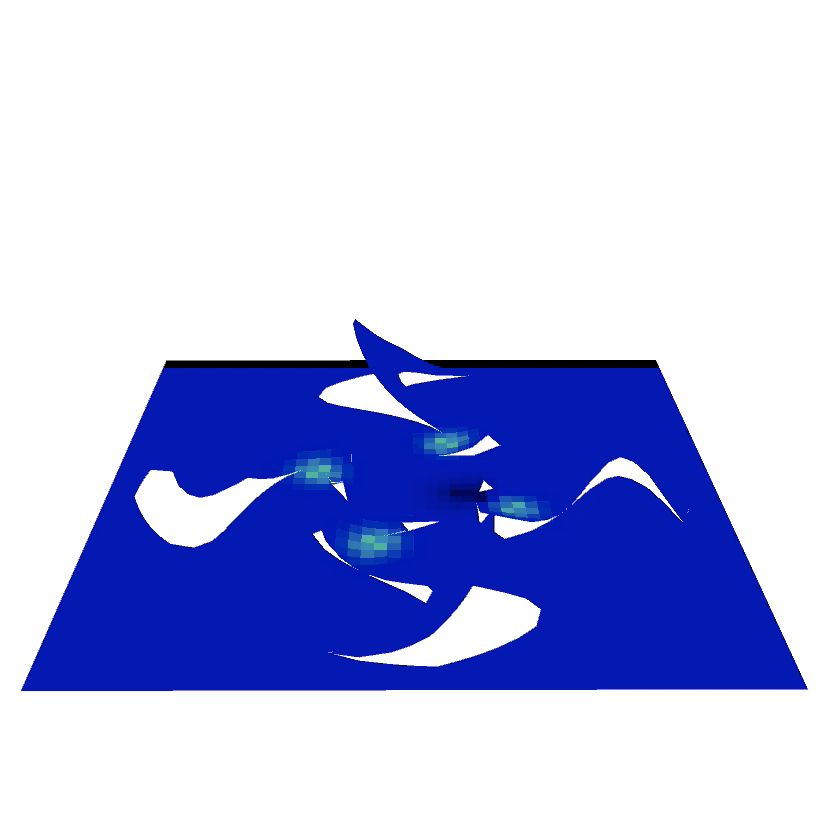}
    \includegraphics[trim=0 0 0 130,clip,width=0.2\textwidth]{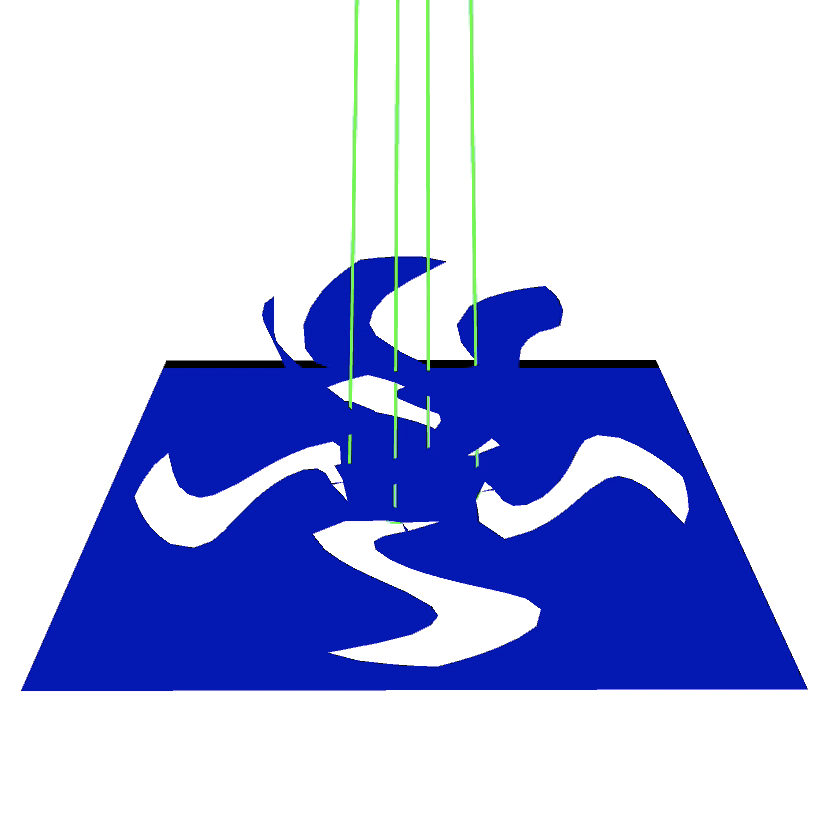} \\
    \includegraphics[trim=0 0 0 130,clip,width=0.2\textwidth]{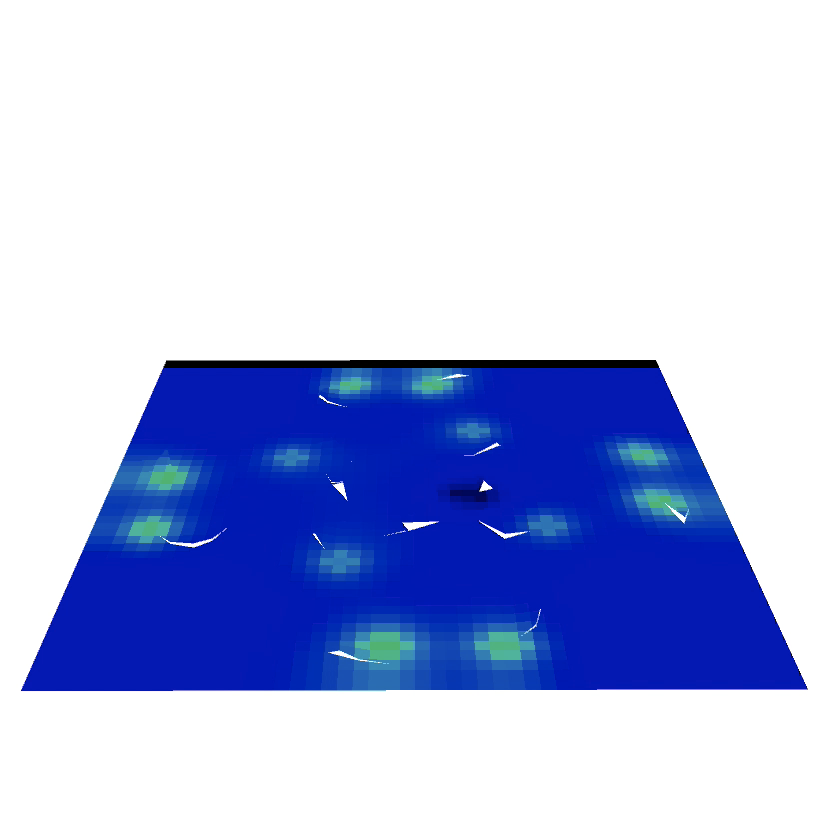}
    \includegraphics[trim=0 0 0 130,clip,width=0.2\textwidth]{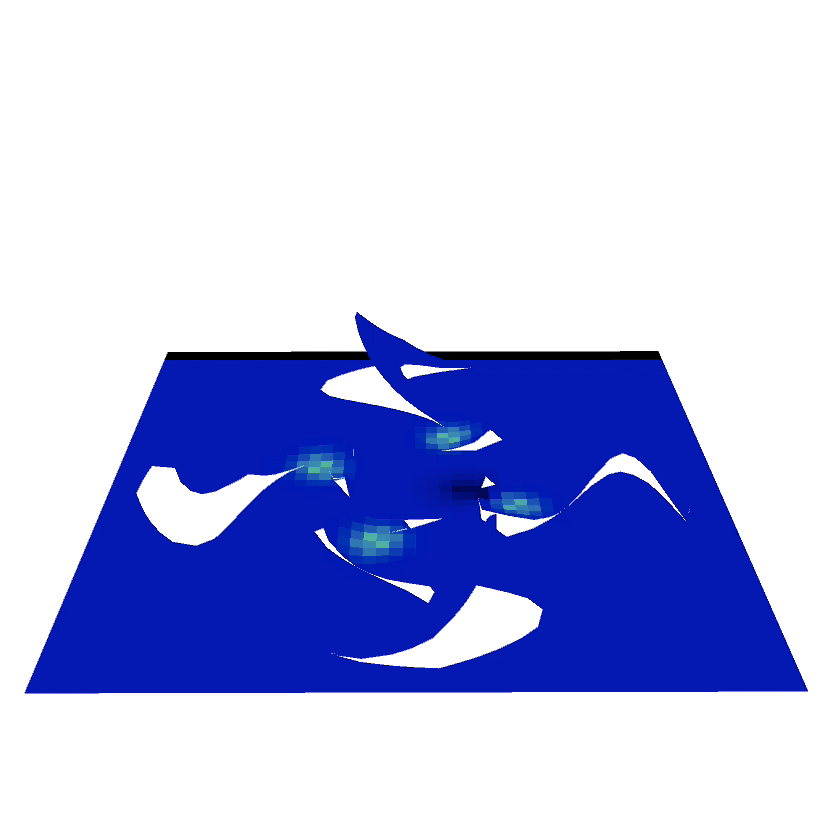}
    \includegraphics[trim=0 0 0 130,clip,width=0.2\textwidth]{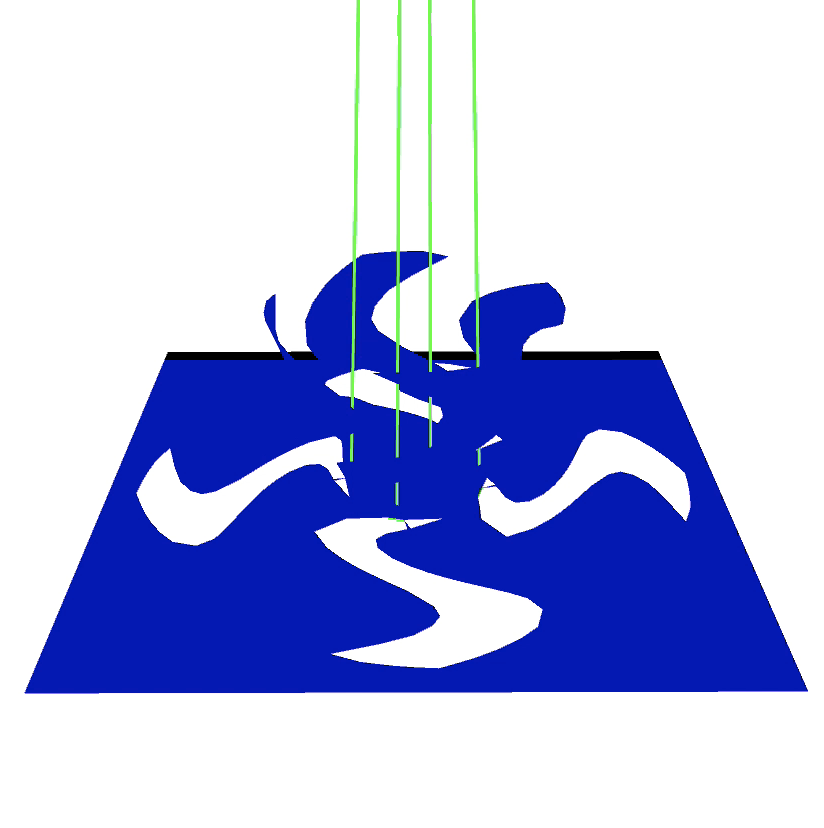}
    \caption{Simulated thermal image for laser forming Sinuous Antenna. Top: overheating due to lack of time delay. Bottom: head minimized strategically using adaptive folding path. }
    \label{fig:sinuous}
\end{figure}

\subsection{Fabrication and Thermal Imaging}
\label{sec:fab}


\begin{figure}[h]
\centering
    \includegraphics[width=0.6\textwidth]{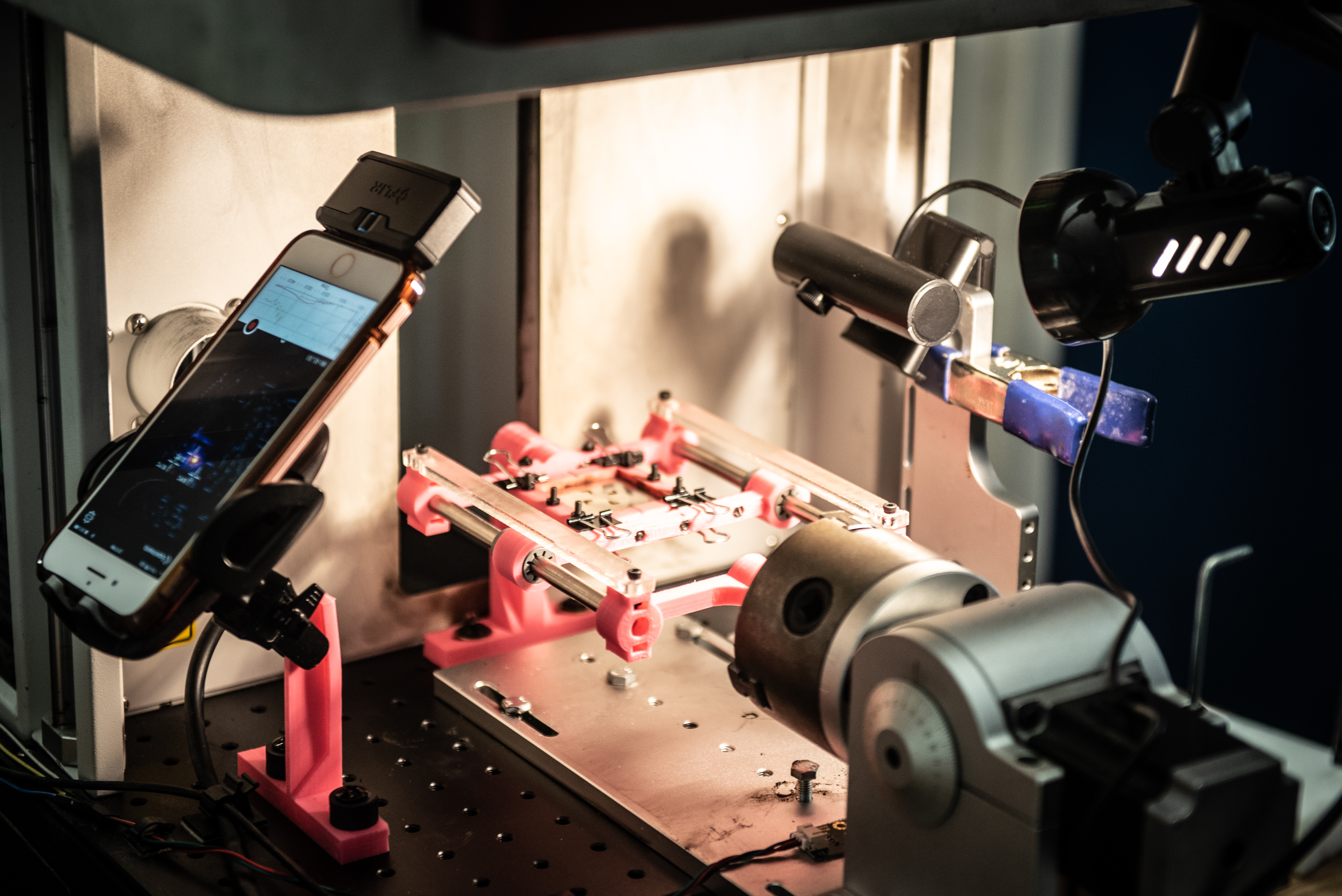}
    \caption{Laser forming origami setup with thermal camera.}
    \label{fig:machine}
\end{figure}
\begin{figure}[h]
\centering
 \includegraphics[trim=0 80 0 80,clip,width=0.8\textwidth]{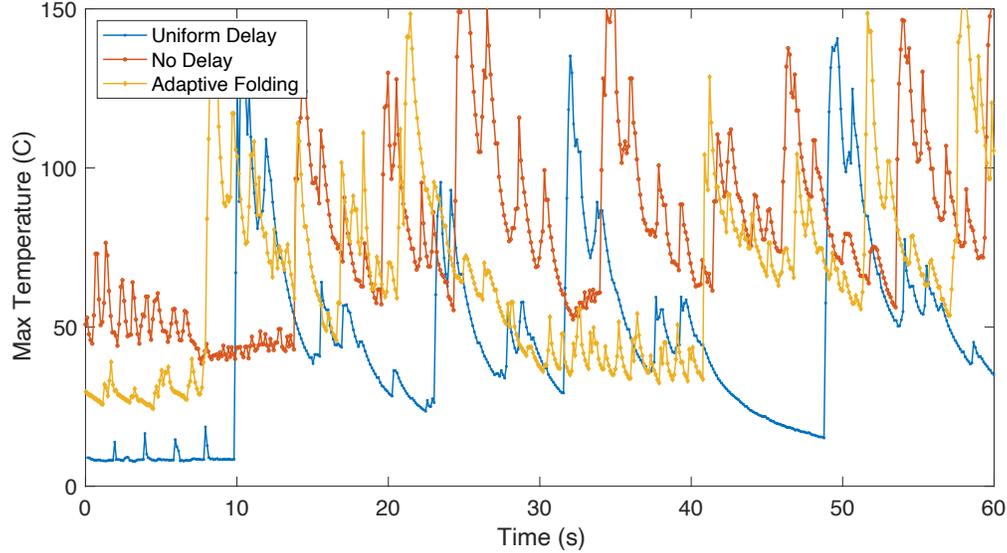}
    \caption{Maximum temperature recorded with the FLIR ONE Pro thermal camera during laser forming of the Sinuous Antenna model. }
    \label{fig:thermal}
\end{figure}

To validate the design in laser forming, we use a 20 W, 1064 nm fiber laser marker and the substrate of 75 um thick stainless steel to fabricate the sinuous antenna model with 190 triangle faces. The laser marker receives the offline folding instructions from the motion planner and fabrication is completely open-loop.
The setup is shown in Fig. \ref{fig:machine}. The fabrication results are shown in Fig. \ref{fig:frame}. 
We also use a thermal camera (FLIR ONE Pro) to capture the thermal image of the process. Fig. \ref{fig:thermal} shows the maximum temperature in the thermal image recorded using the camera in a 60 seconds snapshot. Due to the sensor limitation, the output temperature data are truncated at 150\textcelsius{}. Each of the temperature spike in the chart corresponds to cutting laser strikes to the substrate surface. Comparing to the aggressive no-delay folding and conservative uniform delay motion, our adaptive folding strategically times the process for an efficient fabrication.

\section{CONCLUSIONS}
We presented the idea of the simulation-in-the-loop folding motion planner and demonstrated the benefits that
better estimation of  the foldability can be estimated when material and thermal properties have significant influence on the folding process. 
We used laser forming origami as an example and demonstrated that the proposed approach is a significant improvement over the current  practice that is based on trial-and-error and is inefficient and fragile. Our algorithm takes advantage of the distance from the expansion configuration to the goal configuration as a heuristic to guide the search in the roadmap for the shortest path, which reduces the computational expensive evaluation of the local planning with thermal simulations. 
Future work will explore the integration of high-fidelity simulations, such as those described in Appendix A, that can predict the thermal fields exerted on the sheet by the traveling laser beam with more accuracy, to obtain more physically accurate results from the folding planner.





\appendix

\subsection{High-Fidelity Simulations}


This appendix describes a preliminary method for the performance of high-fidelity finite element (FE) simulations of laser folding and its associated results. 
It is expected in future work to integrate these high-fidelity simulations with the reduced-order thermal model used for path planning to improve the physical accuracy of the planning process.
This appendix describes the process to calibrate the physical model parameters (heat transfer parameters of the laser and material properties of the metal sheet), and the execution of physical simulations of laser folding, which provide insights on the mechanics and heat transfer phenomena involved in this type of folding processes.

\begin{figure}[h]
\centering
    \includegraphics[width=0.6\textwidth]{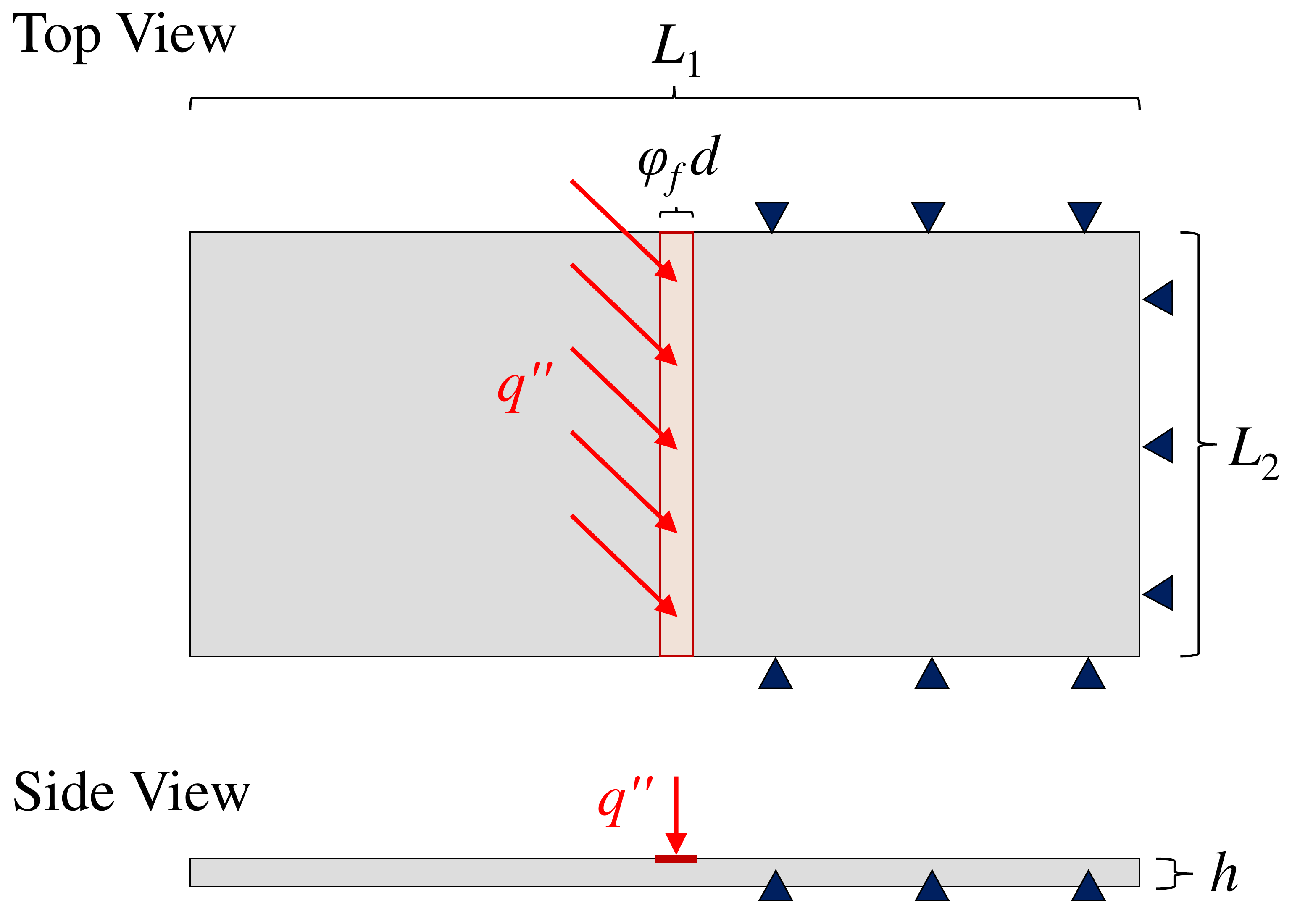}
    \caption{Sheet with a central fold used for calibration of model parameters. 
    The heat flux per unit area applied at the fold is denoted by $q''$. 
    The triangles denoted fixed locations.}
    \label{fig:sheetdomain}
\end{figure}

Initially, fold angle vs.~laser irradiation cycle data for a sheet with a single fold are obtained from experiments.  
Then, a FE model is developed based on the geometry of the experimental sample and material properties provided by the manufacturer of the used metal sheets.  
Calibration parameters are properly adjusted so that the simulation results match the experimental data.
%
%
%
%
%
%
The geometrical domain of the origami sheet with a single central fold employed in the calibration of model parameters is presented in Fig.~\ref{fig:sheetdomain}. 
The sheet has sides with lengths $L_1 = 70$ mm and $L_2 = 35$ mm, and thickness $h = 0.1016$ mm.  
The sheet has a centered fold of length $L_2$ and width $\phi_f d$, where $d$ is the nominal diameter of the laser and $\phi_f$ is a calibration parameter that accounts for the spread of the laser beam diameter when it reaches the fold.  


The room temperature and experimental parameters for the laser are listed in Table~\ref{t:laserparam}.  
The heat flux per unit area $q''$ corresponding to the laser beam irradiation on the sheet is applied to the laser-width region shaded in Fig.~\ref{fig:sheetdomain}.  
The right face of the sheet in Fig.~\ref{fig:sheetdomain} is held fixed. 

\begin{table}[ht!]
\centering
\caption{Thermal parameters used for the simulations.}
\begin{tabular}{ |P{3.5cm}|P{4.0cm}|}
 \hline
Room temperature & 298 K \\ \hline
Laser diameter $d$ (fiber laser) & 0.001064 mm  \\ \hline
Fold power $P_f$ & 2$\times$10$^{9}$ nW (2 W) \\ \hline
Fold speed $v_f$ & 30 mm/s \\ \hline
Delay power & 0 nW (0 W) \\ \hline
Delay speed $v_d$ & 10 mm/s \\ \hline
\end{tabular}
\label{t:laserparam}
\end{table}

The material of the sheet used for the laser folding is M31612-4 Stainless Steel from Maudlin Product.  
The properties for this material are listed in Table~\ref{t:properties}, and are obtained from specifications provided by the manufacturer. 

\begin{table}[ht!]
\centering
\caption{Properties for M31612-4 Stainless Steel.}
\begin{tabular}{ |P{2.6cm}|P{5.2cm}|}
 \hline
Yield strength $\sigma_Y$ & 2.758$\times$10$^{8}$ $\mu$N/mm$^{2}$ (2.758$\times$10$^{8}$ Pa)~\cite{2} \\  \hline 
Density & 0.008 g/mm$^{3}$~\cite{3}
\\ \hline
Young's modulus $E$ & 1.93$\times$10$^{11}$ $\mu$N/mm$^{2}$ (1.93$\times$10$^{11}$ Pa)~\cite{3} \\ \hline
Coefficient of thermal expansion (CTE) 
& 1.75$\times$10$^{-5}$ K$^{-1}$~\cite{4}
\\ \hline
Thermal conductivity 
& 1.63$\times$10$^{7}$ nW/(mm K) (16.3 W/(m K))~\cite{4} \\ \hline
Specific heat 
& 5$\times$10$^{8}$ nJ/(g K) 
(500 J/(kg K))~\cite{3} \\ \hline
\end{tabular}
\label{t:properties}
\end{table}

Table~\ref{t:assumption} provides the material properties that are not reported directly by the manufacturer but are necessary for the simulation process. 
The assumed heat transfer convection coefficient of still air is also provided in the table.
A linear plastic hardening response is assumed, where the first point defining the stress vs.~plastic strain line ($\varepsilon_{p_1}$, $\sigma{p_1}$) corresponds to the yielding point, and the second point ($\varepsilon_{p_2}$, $\sigma{p_2}$) is selected to be at the ultimate strain and stress of the material.

\begin{table}[ht!]
\centering
\caption{Estimated material parameters.}
\begin{tabular}{|P{3.1cm}|P{4.5cm}|}
 \hline
 Plastic strain parameters (isotropic hardening) &
 $\varepsilon_{p_1} = 0$, $\sigma_{p_1} = \sigma_Y$; $\varepsilon_{p_2} = 0.6$, $\sigma_{p_2} =$ 5.516$\times$10$^{8}$ $\mu$N/mm$^{2}$ (5.516$\times$10$^{8}$ Pa)  \\  \hline
 Heat transfer convection coefficient & 10000 nW/(mm$^{2}$ K) (10 W/(m$^2$ K)) \\ \hline
\end{tabular}
\label{t:assumption}
\end{table}


In the laser folding experiment, the laser beam runs across the crease interchangeably between the folding power mode and the delay power mode, with parameters provided in Table~\ref{t:laserparam}.  
Each pass across the crease counts as one cycle, and the laser application continues for the designated number of cycles.  
In the laser folding simulation, the average of total laser power is computed to obtain the heat flux and the execution time to be applied onto the fold region highlighted in Fig.~\ref{fig:sheetdomain} representing the crease for each cycle.
The width of the fold region used in  the simulations corresponds to the nominal laser diameter $d$ multiplied by the fold width calibration factor $\phi_f$. 
The \emph{folding time} $t_f$ corresponds to the time period where each location along the fold region is irradiated by the laser beam.  
It is calculated by dividing the width of the fold region by the speed of the laser beam movement during folding $v_f$ (these parameters are listed in Table~\ref{t:laserparam}): $t_f  = \frac{\phi_f d }{v_f}$. 

The \emph{delay time} $t_d$ is the  time period that each location at the fold region is not irradiated by the laser beam during each cycle in the experiment: $t_d = \frac{L_2}{v_d} + \frac{L_2 - \phi_f d}{v_f}$,
%
%
 where $v_d$ is the speed of the laser beam movement during delays.
%
%
The total folding laser power $P_{f,sum}$ corresponds to the laser folding power $P_f$ multiplied by the ratio of the area of the whole fold region $\phi_f d \times L_2$ to the area of the laser beam region $\phi_f d \times \phi_f d$: $P_{f,sum} = P_f \frac{\phi_f d}{\phi_f d} \frac{L_2}{\phi_f d} = P_f \frac{L_2}{\phi_f d}$.
%
%
%
%
Finally, the heat flux $q''$ applied to the fold region (highlighted in Fig.~\ref{fig:sheetdomain}) is the total folding power $P_{f,sum}$ divided by the  area of the fold region: $q'' = \frac{P_{f,sum}}{L_2 \phi_f d} = \frac{P_{f}}{\phi_f^2 d^2}$.





Abaqus FE analysis software is used to perform the laser folding simulations.  
%
%
Since the 3D shell model has uniform properties and boundary conditions along the fold length direction (of size $L_2$), the simulation can be performed on a small strip along this length direction instead.  
This creates a quasi-plane strain modeling domain.  
\begin{figure}[h]
\centering
    \includegraphics[width=0.6\textwidth]{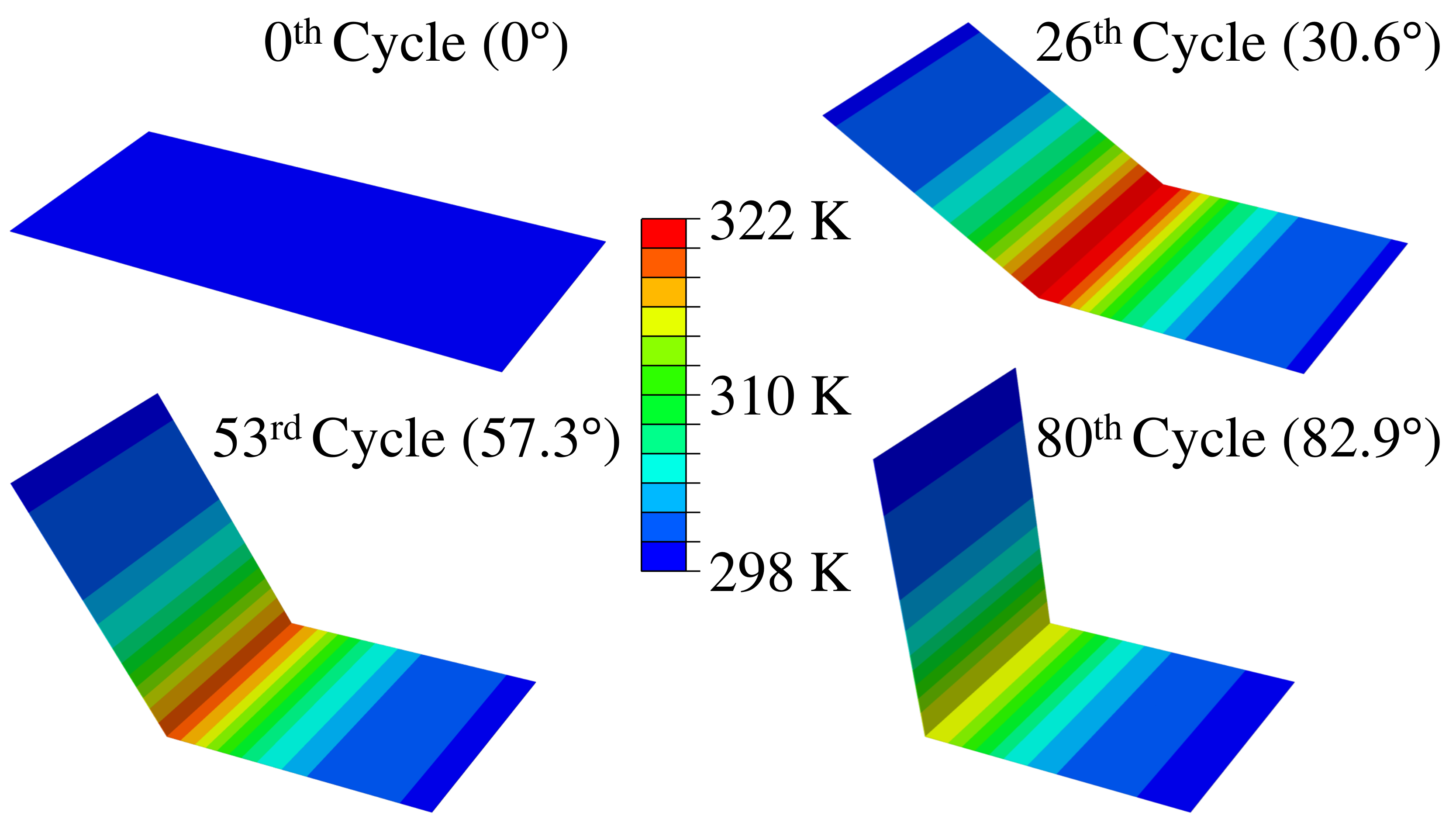}
    \caption{Simulation results for the calibrated model from 0$\degree$ in the 0\textsuperscript{th} cycle to 82.9$\degree$ in the 80\textsuperscript{th} cycle.}
    \label{fig:sim_result}
\end{figure}
%
%
The adjustable parameters assumed for the model calibration are the fold width calibration factor $\phi_f$ and the CTE calibration factor $\phi_\alpha$.  
The CTE used for the simulation corresponds to the value provided by the manufacturer from Table~\ref{t:properties} multiplied by $\phi_\alpha$.  
The fold width factor $\phi_f$ defines the width of the fold region in terms of the laser diameter $d$ as shown in Fig.~\ref{fig:sheetdomain}. 
After testing multiple combinations of calibration parameters (assuming only integers to reduce the amount of test simulations), values of $\phi_f$ of 10 and $\phi_\alpha$ of 2 are found to allow for simulations that accurately represent the experimental data.  
The result of the simulation using the calibrated model is shown in Fig.~\ref{fig:sim_result}. 
Figure~\ref{fig:sim_result} shows  fold angle vs.~laser irradiation cycle from the simulation using the calibrated model and the experimental data.  Good agreement is observed.





\begin{figure}[h]
\centering
    \includegraphics[width=0.6\textwidth]{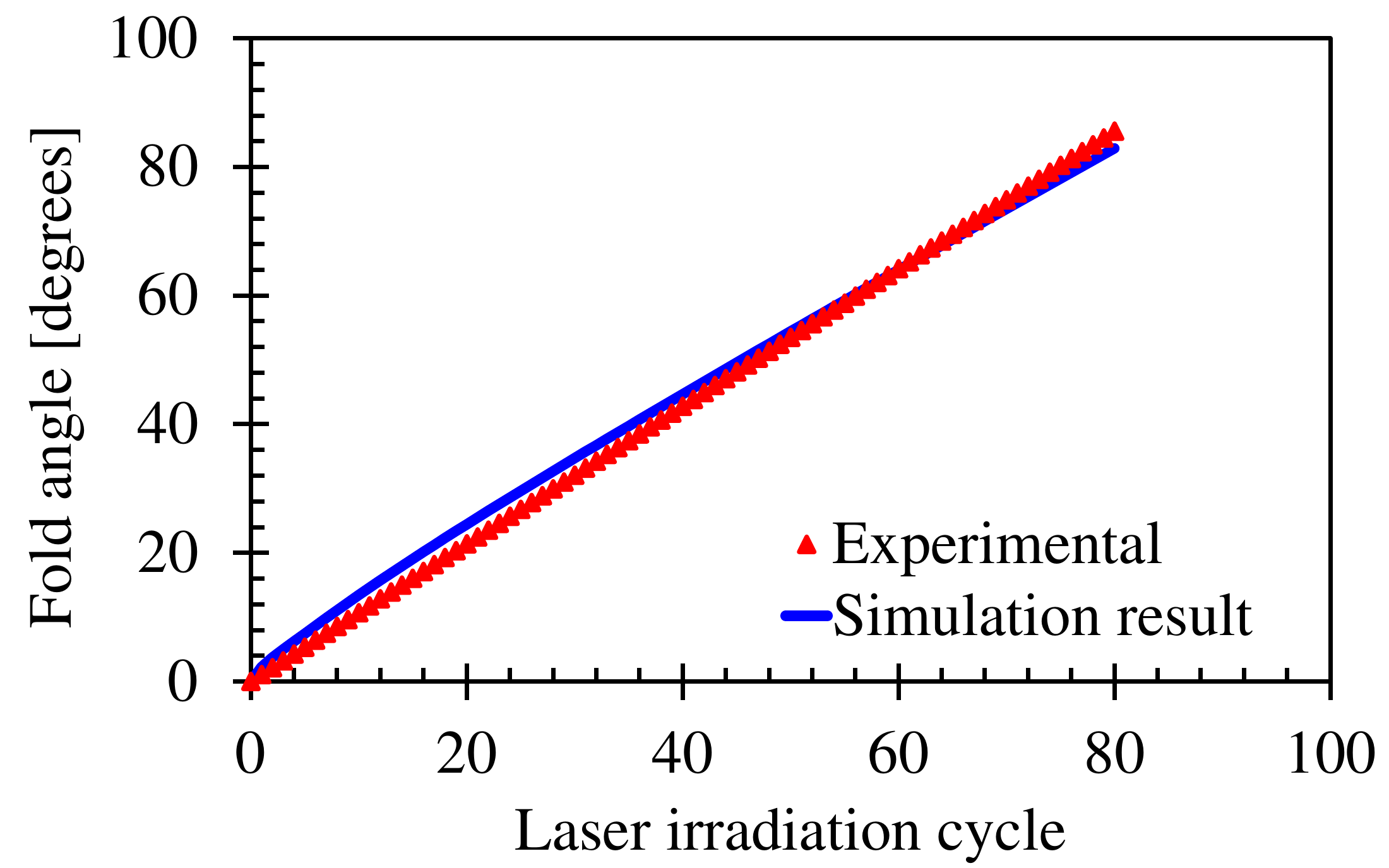}
    \caption{Fold angle vs.~laser irradiation cycle obtained after calibration of $\phi_f$ and $\phi_{\alpha}$ using the experimental data.}
    \label{fig:sim_plot}
\end{figure}
The full temperature field of the sheet is recorded ast each increment during the FE simulation, as displayed in the contour colors of Fig.~\ref{fig:sim_result}.  
In future work, such data obtained from a high-fidelity model will be integrated into the laser path planner to employ more accurate predictions of the thermal field in the sheet.





\section*{Acknowledgments}

W.G.~acknowledges the support 
from the Graduate Assistance in Areas of National Need Program No.~P200A180013 from the U.S.~Department of Education.

\bibliographystyle{IEEEtran}
\bibliography{georobotics,geom,origami,masc,ml,local}

\end{document}